\documentclass{article}

\usepackage{PRIMEarxiv}

\usepackage[utf8]{inputenc} 
\usepackage[T1]{fontenc}    
\usepackage{hyperref}       
\usepackage{url}            
\usepackage{booktabs}       
\usepackage{amsfonts}       
\usepackage{nicefrac}       
\usepackage{microtype}      
\usepackage{lipsum}
\usepackage{fancyhdr}       
\usepackage{graphicx}       
\usepackage{xcolor}
\usepackage[round]{natbib}

\title{A template for Arxiv Style
}

\author{
  Chunchao Ma,  \\
  Department of Computer Science \\
  The University of Sheffield \\
  \texttt{chunchaoma@hotmail.com} \\
   \And
  Arthur Leroy,  \\
  Department of Computer Science \\
  The University of Manchester \\
  \texttt{arthur.leroy.pro@gmail.com} \\
  \And
  Mauricio \'{A}lvarez,  \\
  Department of Computer Science \\
  The University of Manchester \\
  \texttt{mauricio.alvarezlopez@manchester.ac.uk} \\
}

\usepackage{amssymb}
\usepackage{bm}
\usepackage{url}
\usepackage{hyperref}
\hypersetup{
    pdftitle={Latent Variable Multi-output Gaussian Processes for Hierarchical Datasets},
    pdfsubject={},
    pdfauthor={Chunchao Ma, Arthur Leroy, Mauricio \'{A}lvarez},
    pdfkeywords={Multi-output Gaussian processes, Latent variables, Hierarchical data, Variational inference},
    colorlinks = true,
    urlcolor = blue,
    linkcolor = orange,
    citecolor = orange
}
\usepackage{float} 
\usepackage{multirow}
\usepackage{times}
\usepackage{amsmath}

\title{Latent Variable Multi-output Gaussian Processes for Hierarchical Datasets}

\begin{document}
\maketitle

\begin{abstract}
Multi-output Gaussian processes (MOGPs) have been introduced to deal with multiple tasks by exploiting the correlations between different outputs. Generally, MOGPs models assume a flat correlation structure between the outputs. However, such a formulation does not account for more elaborate relationships, for instance, if several replicates were observed for each output (which is a typical setting in biological experiments). This paper proposes an extension of MOGPs for hierarchical datasets (i.e. datasets for which the relationships between observations can be represented within a tree structure). Our model defines a tailored kernel function accounting for hierarchical structures in the data to capture different levels of correlations while leveraging the introduction of latent variables to express the underlying dependencies between outputs through a dedicated kernel. This latter feature is expected to significantly improve scalability as the number of tasks increases. An extensive experimental study involving both synthetic and real-world data from genomics and motion capture is proposed to support our claims.
\end{abstract}




\section{Introduction}
\label{Introduction}
In Bayesian statistics, hierarchical designs are a way to represent generative models that take multi-level structures of correlation into consideration.
A hierarchical dataset can generally be represented as a top-down tree-like architecture. We refer to all leaf nodes of the same level as replicas since they inherit from the same parent node. The authors of \cite{kalinka2010gene} proposed a dataset, which we used in our experiments, where gene expression is observed through eight replicas. Gene expression is a biological process indicating how the information of a particular gene can affect the phenotype, and many practitioners aim to understand this phenomenon better. In real-world applications, many datasets present a hierarchical structure, such as the one observed in this gene expression dataset.  

In a hierarchical model, prior distributions of the parameters of interest generally depend upon other parameters (often called hyper-parameters) that also have their own prior distribution \citep{gelman2013bayesian}. Standard \emph{flat} (i.e. non-hierarchical) modelling strategies often struggle to fit hierarchical datasets adequately with a reasonable number of parameters. Conversely, they can be prone to overfitting as the number of parameters increases \citep{gelman2013bayesian}. However, those issues can be avoided when properly designing the hierarchical structure in modelling assumptions. In our previous example, a model designed with a hierarchical structure appears as a natural choice to account for correlations between leaf nodes (or replicas).

In the Gaussian processes (GP) literature, the topic of hierarchical modelling has quickly emerged as a promising approach to tackle a wide range of problems. More specifically, \cite{lawrence2007hierarchical} first introduced a hierarchical Gaussian process model for dimensionality reduction. Then, the two-layer hierarchical approximation proposed in \cite{park2010hierarchical} helped to reduce the computational complexity of standard GP regression. Later, \cite{hensman2013hierarchical} derived a novel hierarchical kernel to handle gene expression data, while \cite{damianou2013deep} established a deep-layer model where each layer was based on a Gaussian processes mapping. The paper \cite{flaxman2015fast} also developed a hierarchical model through a prior distribution over kernel hyperparameters and used MCMC for inference. More recently, \cite{li2018hierarchical} proposed a hierarchical formulation extracting latent features from the input dataset through the GP latent variable model and derived a Bayesian inference procedure to generate outputs based on those latent features. 

None of the aforementioned models is yet adapted to the case of multiple-output GPs, where each output presents an underlying hierarchical structure. In this sense, previous models would generally fail to capture the correlation existing between each replica. Moreover, to the best of our knowledge, no method is currently able to predict entirely missing replicas. This paper aims to fill this gap by providing an extension of the latent variable multi-output Gaussian process (LVMOGP) model \citep{dai2017efficient} that can cope with hierarchical datasets and naturally predict missing replicas. Interestingly, our model could also be viewed as a generalisation of hierarchical GPs (HGP) \citep{hensman2013hierarchical}, as it somewhat combines the two approaches. Therefore, we named this method \emph{hierarchical multi-output Gaussian processes with latent variables} (HMOGP-LV). More specifically, \textbf{HMOGP-LV} controls the correlation between outputs through latent variables and captures the structure of data using a hierarchical kernel. Using inducing variables that share information of all replicas across the outputs, our model can predict missing points and entirely missing replicas. In this sense, our model tackles a more general problem, which the standard HGP model did not handle. When predicting a missing replica from one output, the inducing variables can use information from the corresponding replicas in other outputs. We derived an analytical approximation scheme for \textbf{HMOGP-LV} in two different settings: all outputs having the same input data; all outputs having specific input data.

\section{Model and assumptions}
In this section, let us formally derive the hierarchical multi-output Gaussian processes with latent variables (HMOGP-LV). We first present HMOGP-LV in a setting where all outputs are observed on the same input set. Further, the model is extended to deal with cases where each output has its own input set.

\subsection{Hierarchical Multi-output Gaussian Processes with Latent Variables}
\label{HMOGPLV: sameinput}

Assume that we observe a $D$-dimensional output vector $\mathbf{y}(\mathbf{x})=\Big[\mathbf{y}_{1}^{\top}(\mathbf{x}),$ $\mathbf{y}_{2}^{\top}(\mathbf{x}),$ $\cdots,$ $\mathbf{y}_{D}^{\top}(\mathbf{x})\Big]^{\top}$, where $\mathbf{x} \in \mathbf{R}^{v}$ is the input vector (of an arbitrary dimension $v$). To encode the hierarchical structure of the data, we assume that $R$ replicas are observed for each output. Therefore, for all $d = 1, \dots, D$, each component can be decomposed as $\mathbf{y}_{d}(\mathbf{x})=\left[y_{d}^{1}(\mathbf{x}), y_{d}^{2}(\mathbf{x}), \cdots, y_{d}^{R}(\mathbf{x})\right]^{\top}$, where $y_{d}^{r}(\mathbf{x})$ is the $r$-th replica of the $d$-th output evaluated at $\mathbf{x}$. 
For the sake of simplicity, we assume that each replica presents the same number $N$ of data points (although the following would still hold otherwise, up to minor technical adjustments). 
Formally, each replica $y_{d}^{r}(\mathbf{x})$ could be modelled as a latent random function $f_{d}^{r}(\mathbf{x})$ corrupted by a Gaussian white noise $\epsilon_{d}$ with $\sigma_{d}^2$ variance:
    \begin{align}
    y_{d}^{r}\left(\mathbf{x}\right) &=f_{d}^{r}\left(\mathbf{x}\right)+\epsilon_{d} \\
    \quad f_{d}^{r}(\mathbf{x}) &\sim \mathcal{G}\mathcal{P}\left(0, k_{f}\left(\mathbf{x},
    \mathbf{x}^{\prime}\right)\right) \\
    \quad \epsilon_{d} &\sim \mathcal{N}\left(0, \sigma_{d}^2 \right).
    \end{align}
We refer to the collection of the $r$-th observed input data points as $\mathbf{X}_{r} = [\mathbf{x}_{r}^{(1)},\cdots,$ $\mathbf{x}_{r}^{(N)}]^\top \in \mathbf{R}^{N \times v}$, and to the associated outputs as $\mathbf{y}_{d}^{r} = [y_{d}^{r}\left(\mathbf{x}_{r}^{(1)}\right),$ $\cdots, y_{d}^{r}\left(\mathbf{x}_{r}^{(N)}\right)]^\top$ $\in \mathbf{R}^{N}$ for the $r$-th replica of the $d$-th output. The $d$-th input and output sets are denoted $\mathbf{X}=\left\{\mathbf{X}_{r}\right\}_{r=1}^{R}$ and $\mathbf{y}_{d}=\left\{\mathbf{y}_{d}^r\right\}_{r=1}^{R}$, respectively. Finally, the vector $\mathbf{y} = [\mathbf{y}_{1}^\top, \cdots, \mathbf{y}_{D}^\top]^\top$ refers to all observed outputs.

To cope with the assumed hierarchical structure, we still need to define an additional layer of correlation in the generative model. Therefore, suppose that an underlying function controls the mean parameter of the prior distribution from which the replicas are drawn. Let us denote this function as $g(\cdot)$, a zero mean GP with covariance $k_{g}\left(\cdot, \cdot \right)$ such as $g(\mathbf{x}) \sim \mathcal{G} \mathcal{P}\left(0, k_{g}\left(\mathbf{x}, \mathbf{x}^{\prime}\right)\right)$. Similarly to the hierarchical structure proposed in \cite{hensman2013hierarchical}, all latent functions are assumed to be drawn from a Gaussian process with a $g(\mathbf{x})$ mean and a $k_{f}\left(\mathbf{x}, \mathbf{x}^{\prime}\right)$ covariance. Overall, we obtain:

\begin{align}
g(\mathbf{x}) & \sim \mathcal{G} \mathcal{P}\left(0, k_{g}\left(\mathbf{x}, \mathbf{x}^{\prime}\right)\right), \\
f_{d}^{r}(\mathbf{x}) & \sim \mathcal{G} \mathcal{P}\left(g(\mathbf{x}), k_{f}\left(\mathbf{x}, \mathbf{x}^{\prime}\right)\right), \\
y_{d}^{r}\left(\mathbf{x}\right) & =f_{d}^{r}\left(\mathbf{x}\right)+\epsilon_{d}.
\label{y_d}
\end{align}

Intuitively, the above generative model indicates that all outputs share information both through kernel functions $k_{g}\left(\cdot, \cdot\right)$ and $k_{f}\left(\cdot, \cdot \right)$. 

In order to replace the fixed coregionalisation matrix with a kernel matrix, we now assume there exists a continuous latent vector $\mathbf{h}_{d}\in \mathbf{R}^{Q_{H}}$ associated with each output $\mathbf{y}_{d}$. $Q_{H}$ is set in advance by the modeller. From a learning point of view, the latent variables are ultimately extracted from observations by maximising the marginal likelihood. Latent variables of all outputs are stacked into $\mathbf{H}=\left[\mathbf{h}_{1}^{\top}, \ldots, \mathbf{h}_{D}^{\top}\right]^{\top}$ and each of them follows the same prior distribution (e.g. a normal distribution). Therefore, we now obtain the following: 

\begin{align}
g(\mathbf{x}) & \sim \mathcal{G} \mathcal{P}\left(0, k_{g}\left(\mathbf{x}, \mathbf{x}^{\prime}\right)\right), \\
f_{d}^{r}(\mathbf{x})  & \sim \mathcal{G} \mathcal{P}\left(g(\mathbf{x}), k_{f}\left(\mathbf{x}, \mathbf{x}^{\prime}\right)\right), \\
y_{d}^{r}(\mathbf{x}) & =f_{d}^{r}\left(\mathbf{x}, \mathbf{h}_{d}\right)+\epsilon_{d}, \ \mathbf{h}_{d} \sim \mathcal{N}(\mathbf{0}, \mathbf{I}).
\label{HMOGPLV: y_f_latent}
\end{align}

\begin{figure}
\centering
\includegraphics[width=0.8\textwidth]{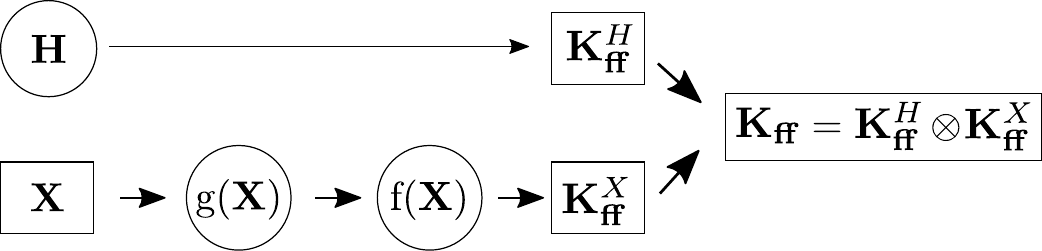}
\caption{Summary of the generative procedure used to derive the overall covariance structure. $\mathbf{K}_{\mathbf{f} \mathbf{f}}^{X}$ contains the hierarchical structure of our model; $\mathbf{K}_{\mathbf{f} \mathbf{f}}^{H}$ contains the correlation between each output.}
\label{Kernel_final}
\end{figure}

There are many ways to build our kernel based on Eq. \eqref{HMOGPLV: y_f_latent}. 
The overall kernel matrix is built through a Kronecker product to account for all correlations between inputs and outputs, as illustrated in Figure \ref{Kernel_final}.
We first build a kernel matrix for the outputs:

\begin{align}
\mathbf{K}_{\mathbf{f} \mathbf{f}}^{H} = \left(\begin{array}{ccc}K^{H}_{1,1}  & \ldots & K^{H}_{1,D}  \\\vdots & \ddots & \vdots \\K^{H}_{D,1} & \ldots & K^{H}_{D,D}\end{array}\right),
\end{align}

where $K^{H}_{i,j} =
k_{H}\left(\mathbf{h}_{i}, \mathbf{h}_{j}\right)$ describes the correlation between $i$-th and $j$-th outputs and $k_{H}(\cdot,\cdot)$ is a kernel function. Compared with a fixed coregionalisation matrix, $k_{H}$ is still able to produce flexible matrices while dramatically reducing computational complexity in high dimensional applications. By leveraging $k_{H}$ and latent variables $\mathbf{H}$, this approach has previously demonstrated efficiency in avoiding over-fitting \citep{dai2017efficient} and dealing with scarce data sets.

Let us now derive a kernel matrix over the inputs. 
Since there exists a linear hierarchical structure for our latent functions, if two input points are associated with the same output and $r$-th replica (e.g., $\mathbf{x}_{r}^{(i)}$ and $\mathbf{x}_{r}^{(j)}$), the corresponding GP distribution is characterised by a compound covariance function $k_{g}^{f} \left(\mathbf{x}_{r}^{(i)}, \mathbf{x}_{r}^{(j)}\right) = k_{f}\left(\mathbf{x}_{r}^{(i)}, \mathbf{x}_{r}^{(j)}\right) + k_{g}\left(\mathbf{x}_{r}^{(i)}, \mathbf{x}_{r}^{(j)}\right)$. 
Conversely, for input points coming from different replicas, the covariance structure becomes $k_{g}\left(\mathbf{x}_{r}^{(i)}, \mathbf{x}_{r^{\prime}}^{(j)}\right)$. 
We denote $k_{\text{h}}\left(\cdot, \cdot \right)$ (where the index \emph{h} stands for \emph{hierarchy}) the kernel function defined as:

\begin{equation}
k_{\text{h}}\left(\mathbf{x}_{r}^{(i)}, \mathbf{x}_{r^{\prime}}^{(j)} \right) = \left\{\begin{aligned}
k_{g}^f \left(\mathbf{x}_{r}^{(i)}, \mathbf{x}_{r^{\prime}}^{(j)}\right), \ \text{$r = r^{\prime}$} \\
k_{g}\left(\mathbf{x}_{r}^{(i)}, \mathbf{x}_{r^{\prime}}^{(j)} \right), \ \text {$r \ne r^{\prime}$}
\end{aligned}\right.
\end{equation}

where $\mathbf{x}_{r}^{(i)}, \mathbf{x}_{r^{\prime}}^{(j)} \in \mathbf{X}$. 
The covariance matrix $\mathbf{K}_{\mathbf{f} \mathbf{f}}^{X}$ obtained by evaluating this hierarchical kernel on input points can be expressed as:

\begin{align} 
\mathbf{K}_{\mathbf{f} \mathbf{f}}^{X} = \left(\begin{array}{ccc} k_{g}^f \left(\mathbf{X}_{1}, \mathbf{X}_{1}\right) & \ldots & k_{g}\left(\mathbf{X}_{1}, \mathbf{X}_{R}\right) \\\vdots & \ddots & \vdots \\ k_{g}\left(\mathbf{X}_{R}, \mathbf{X}_{1}\right) & \ldots &  k_{g}^f \left(\mathbf{X}_{R}, \mathbf{X}_{R}\right) \end{array}\right).
\label{HMOGPLV:KX}
\end{align}

Finally, the covariance matrix of our proposed model is defined as

\begin{align}
\mathbf{K}_{\mathbf{f} \mathbf{f}} = \mathbf{K}_{\mathbf{f} \mathbf{f}}^{H} \otimes \mathbf{K}_{\mathbf{f} \mathbf{f}}^{X},
\label{HMOGPLV: Kernel_HMOGPLV}
\end{align}

where $\otimes$ denotes the Kronecker product between matrices.
Based on Eq. \eqref{HMOGPLV: Kernel_HMOGPLV}, we can derive the prior distribution of $\mathbf{f}=\left[\mathbf{f}_{1}^\top, \ldots, \mathbf{f}_{D}^\top\right]^{\top}$ and the conditional likelihood:

\begin{align}
p\left(\mathbf{f} \mid \mathbf{X}, \mathbf{H}\right)=\mathcal{N}\left(\mathbf{f} \mid \mathbf{0}, \mathbf{K}_{\mathbf{f} \mathbf{f}} \right), \\
p\left(\mathbf{y} \mid \mathbf{X}, \mathbf{f}, \mathbf{H} \right) = \mathcal{N}\left(\mathbf{y} \mid \mathbf{f}, \boldsymbol{\Sigma} \right),
\label{HMOGPLV: latent_distribution}
\end{align}

where $\boldsymbol{\Sigma} \in \mathbf{R}^{NRD \times NRD}$ is a diagonal matrix with a noise variance that can depend on both the particular output $d$ and the particular replica $r$. Thus, the corresponding marginal likelihood can be expressed as (while omitting conditioning on $\mathbf{X}$ for clarity):

\begin{equation}
p\left(\mathbf{y}  \right) = \int p\left(\mathbf{y} \mid \mathbf{f}, \mathbf{H} \right) p\left(\mathbf{f} \mid \mathbf{H}\right)  p\left(\mathbf{H}\right)\text{d}\mathbf{f}\text{d}\mathbf{H}.
\label{ch5:marginallikelihoodhf}
\end{equation}

\subsection{Extension for Different Sets of Inputs}

In the above section, we derived a model that deals with multiple outputs sharing the same input set.
However, in real-world applications, each output may often be observed at different locations. 
In this context, the $d$-th input with replicated data is expressed as $\mathbf{X}_{d}=\left\{\mathbf{X}_{d,r}\right\}_{r=1}^{R}$, where $\mathbf{X}_{d,r} = [\mathbf{x}_{d,r}^{(1)}, \cdots, \mathbf{x}_{d,r}^{(N_{d})}]^\top$. 
Although the general model formulation described in Section \ref{HMOGPLV: sameinput} is preserved, we now need to take extra care when dealing with missing data. 
The specific equations associated with the learning procedure in this framework are detailed in Section \ref{inference:diffinputs}.

\section{Inference}
In general, the integral in the marginal likelihood expression (\ref{ch5:marginallikelihoodhf}) is intractable.
Therefore, we must resort to a variational approximation scheme by deriving a lower bound of the log marginal likelihood.
Our method can also deal with large-scale datasets based on similar ideas and notation, as in \cite{dai2017efficient}.

\subsection{Scalable Variational Inference}

Let us first introduce inducing variables $\mathbf{U} \in \mathbf{R}^{M_{\mathbf{X}} \times M_{\mathbf{H}}}$ associated with our previous outputs and $\mathbf{U}_{:}=\operatorname{vec}(\mathbf{U})$, where ``:'' denotes the vectorisation of a matrix.
We assume that the prior distribution of $\mathbf{U}_{:}$ can be expressed as $p\left(\mathbf{U}_{:}\right)=\mathcal{N}\left(\mathbf{U}_{:} \mid \mathbf{0}, \mathbf{K}_{\mathbf{U} \mathbf{U}}\right)$.
In particular, $\mathbf{K}_{\mathbf{U} \mathbf{U}}$ is supposed to have a similar format as Eq. \eqref{HMOGPLV: Kernel_HMOGPLV}: $\mathbf{K}_{\mathbf{U} \mathbf{U}}=\mathbf{K}_{\mathbf{U} \mathbf{U}}^{H}  \otimes \mathbf{K}_{\mathbf{U} \mathbf{U}}^{X}$. The matrix $\mathbf{K}_{\mathbf{U} \mathbf{U}}^{H}$ is obtained by evaluating $k_{H}(\cdot,\cdot)$ on the inducing outputs $\mathbf{Z}^{H}=\left[\mathbf{z}_{1}^{H}, \ldots, \mathbf{z}_{M_{\mathbf{H}}}^{H}\right]^{\top}$, $\mathbf{z}_{m}^{H} \in \mathbf{R}^{Q_{H}}$.

Similarly, $\mathbf{K}_{\mathbf{U} \mathbf{U}}^{X}$ can be computed with the kernel function $k_{\text{h}}(\cdot,\cdot)$ evaluated on inducing input locations $\mathbf{Z}^{X}$ where $\mathbf{Z}^{X}= \{\mathbf{Z}^{X}_{r}\}_{r=1}^{R}$. $\mathbf{Z}^{X}_{r}$ corresponds with the $r$-th replica and $\mathbf{Z}^{X}_{r}=$ $\left[\mathbf{z}_{r,1}^{X}, \ldots, \mathbf{z}_{r,M_{r}}^{X}\right]^{\top}$ in which $\mathbf{z}_{r,m}^{X} \in \mathbf{R}^{v}$, and $M_{r}$ is the number of inducing input points in the $r$-th replica and $M_{\mathbf{X}} = M_{r} \times R$.
Similar to the inducing variables framework in \cite{titsias2009variational}, the conditional distribution of $\mathbf{f}$ can be expressed as (the inputs $\mathbf{Z}^{X}, \mathbf{Z}^{H}, \mathbf{X}$ and $\mathbf{H}$ are omitted in conditioning for clarity):

\begin{equation}
p\left(\mathbf{f} \mid \mathbf{U}\right)=\mathcal{N}\left(\mathbf{f} \mid \mathbf{K}_{\mathbf{f} \mathbf{U}} \mathbf{K}_{\mathbf{U} \mathbf{U}}^{-1} \mathbf{U}_{:}, \mathbf{K}_{\mathbf{f} \mathbf{f}}-\mathbf{K}_{\mathbf{f} \mathbf{U}} \mathbf{K}_{\mathbf{U} \mathbf{U}}^{-1} \mathbf{K}_{\mathbf{f} \mathbf{U}}^{\top}\right),
\label{sparsegp}
\end{equation}

where $\mathbf{K}_{\mathbf{f} \mathbf{U}}=\mathbf{K}_{\mathbf{f} \mathbf{U}}^{H} \otimes \mathbf{K}_{\mathbf{f} \mathbf{U}}^{X}$. $\mathbf{K}_{\mathbf{f} \mathbf{U}}^{X}$ denotes the cross-covariance matrix computed by evaluating $k_{\text{h}}(\cdot,\cdot)$
between $\mathbf{X}$ and $\mathbf{Z}^{X}$; $\mathbf{K}_{\mathbf{f} \mathbf{U}}^{H}$ is the cross-covariance computed between $\mathbf{H}$ and $\mathbf{Z}^{H}$ with $k_{H}(\cdot,\cdot)$. The underlying graphical models summarising the different assumptions on the kernel structures are displayed in Figure \ref{Kernel_final_inducing_variables} of the Appendix.
As for covariance matrix \eqref{HMOGPLV:KX}, we can define:

\begin{align}
\mathbf{K}_{\mathbf{U}\mathbf{U}}^{X} = \left(\begin{array}{ccc} k_{g}^f\left(\mathbf{Z}^{X}_{1}, \mathbf{Z}^{X}_{1}\right)  & \ldots & k_{g}\left(\mathbf{Z}^{X}_{1}, \mathbf{Z}^{X}_{R}\right) \\\vdots & \ddots & \vdots \\ k_{g}\left(\mathbf{Z}^{X}_{R}, \mathbf{Z}^{X}_{1}\right) & \ldots & k_{g}^f\left(\mathbf{Z}^{X}_{R}, \mathbf{Z}^{X}_{R}\right) \end{array}\right),
\end{align}

and

\begin{align}
\mathbf{K}_{\mathbf{f} \mathbf{U}}^{X}  = \left(\begin{array}{ccc} k_{g}^f\left(\mathbf{X}_{1}, \mathbf{Z}^{X}_{1}\right)  & \ldots & k_{g}\left(\mathbf{X}_{1}, \mathbf{Z}^{X}_{R}\right) \\\vdots & \ddots & \vdots \\k_{g}\left(\mathbf{X}_{R}, \mathbf{Z}^{X}_{1}\right) & \ldots &  k_{g}^f\left(\mathbf{X}_{R}, \mathbf{Z}^{X}_{R}\right) \end{array}\right).
\end{align}

To approximate posteriors over $\mathbf{f}$ and $\mathbf{H}$, we derive a variational distribution $q(\mathbf{f},$ $\mathbf{U}_{:}, \mathbf{H})$ $=p(\mathbf{f} \mid \mathbf{U}_{:}, \mathbf{H}) q(\mathbf{U}_{:})q(\mathbf{H})$. To compute optimal parameters and hyperparameters for our model, we can maximise the associated lower bound of $\log p(\mathbf{y})$ (see Sections \ref{appndx:HMOGPLV-ELBO} and \ref{appndx:HMOGPLV-ExpectationFindsameoutput} of the Appendix for technical details):

\begin{equation}
\mathcal{L} = \mathcal{F}-\operatorname{KL}(q(\mathbf{U}_{:}) \| p(\mathbf{U}_{:}))-\operatorname{KL}(q(\mathbf{H}) \| p(\mathbf{H})),
\label{elbo}
\end{equation}

where we assume $q(\mathbf{U}_{:})=\mathcal{N}\left(\mathbf{U}_{:} \mid \mathbf{M}_{:}, \bm{\Sigma}^{\mathbf{U}_{:}}\right)$ with $\mathbf{M}_{:}$ and $\bm{\Sigma}^{\mathbf{U}_{:}}$ being variational parameters, and

\begin{align} \mathcal{F}=  &-\frac{DRN}{2} \log 2 \pi \sigma^{2}-\frac{1}{2 \sigma^{2}} \mathbf{y}^{\top} \mathbf{y} +\frac{1}{\sigma^{2}} \mathbf{y}^{\top} \Psi \mathbf{K}_{\mathbf{U} \mathbf{U}}^{-1} \mathbf{M}_{:} \nonumber \\ & -\frac{1}{2 \sigma^{2}}  \text{Tr}\left(\mathbf{K}_{\mathbf{U} \mathbf{U}}^{-1} \Phi \mathbf{K}_{\mathbf{U} \mathbf{U}}^{-1}\left(\mathbf{M}_{:} \mathbf{M}_{:}^{\top}+\bm{\Sigma}^{\mathbf{U}_{:}}\right)\right) \nonumber \\ &-\frac{1}{2 \sigma^{2}}\left(\text{Tr}\left\langle\mathbf{K}_{\mathbf{f} \mathbf{f}}\right\rangle_{q\left(\mathbf{H}\right)}-\text{Tr}\left(\mathbf{K}_{\mathbf{U} \mathbf{U}}^{-1} \Phi\right)\right),
\label{sameformula}
\end{align}

where $\Phi  = \left\langle\mathbf{K}_{\mathbf{f}  \mathbf{U}}^{\top}\mathbf{K}_{\mathbf{f}  \mathbf{U}}\right\rangle_{q(\mathbf{H})}$ and $\Psi =\left\langle\mathbf{K}_{\mathbf{f}  \mathbf{U}}\right\rangle_{q(\mathbf{H})}$. 

Notice that the computational complexity of the lower bound is dominated by the product $\mathbf{K}_{\mathbf{f}  \mathbf{U}}^{\top}\mathbf{K}_{\mathbf{f}  \mathbf{U}}$ that is $\mathcal{O}\left(NDRM_{\mathbf{X}}^{2}M_{\mathbf{H}}^{2}\right).$

\subsection{Lower Bound for Different Sets of Inputs}
\label{inference:diffinputs}

When the input locations differ among outputs, the expression in \eqref{elbo} still holds for the lower bound of the log-marginal likelihood. 
However, the term $\mathcal{F}$ needs to be reformulated as (see Section \ref{appndx:HMOGPLV-Different_Outputs} of the Appendix for technical details):
\begin{align} 
\mathcal{F}=& \sum_{d=1}^{D} -\frac{N_{d}R}{2} \log 2 \pi \sigma_{d}^{2}-\frac{1}{2 \sigma_{d}^{2}} \mathbf{y}_{d}^{\top} \mathbf{y}_{d} \nonumber \\  
&+  \frac{1}{\sigma_{d}^{2}} \mathbf{y}_{d}^{\top} \Psi_{d}\mathbf{K}_{\mathbf{U} \mathbf{U}}^{-1} \mathbf{M}_{:} - \frac{1}{2 \sigma_{d}^{2}} \left( \psi_{d} - \text{Tr}\left[\mathbf{K}_{\mathbf{U} \mathbf{U}}^{-1} \Phi_{d}\right] \right) \nonumber \\ 
&- \frac{1}{2\sigma_{d}^{2}} \text{Tr} \left[\mathbf{K}_{\mathbf{U} \mathbf{U}}^{-1} \Phi_{d}\mathbf{K}_{\mathbf{U} \mathbf{U}}^{-1}\left(\mathbf{M}_{:}\mathbf{M}_{:}^{\top} + \bm{\Sigma}^{\mathbf{U}_{:}} \right)\right],
\label{Differentinput:F}
\end{align}

where $\Phi_{d} = \left\langle\mathbf{K}_{\mathbf{f}_{d}  \mathbf{U}}^{\top}\mathbf{K}_{\mathbf{f}_{d}   \mathbf{U}}\right\rangle_{q(\mathbf{h}_{d})}$, $\Psi_{d} =\left\langle\mathbf{K}_{\mathbf{f}_{d}  \mathbf{U}} \right\rangle_{q(\mathbf{h}_{d})}$ and $\psi_{d} =\text{Tr}\left\langle\mathbf{K}_{\mathbf{f}_{d} \mathbf{f}_{d}}\right\rangle_{q\left(\mathbf{h}_{d}\right)}$.

Interestingly, the two KL divergence terms in \eqref{elbo} remain identical in both cases, as they do not depend on the data. 
The product $\mathbf{K}_{\mathbf{f}_{d}  \mathbf{U}}^{\top}\mathbf{K}_{\mathbf{f}_{d}  \mathbf{U}}$ now drives the $\mathcal{O}(N_{d}RM_{\mathbf{X}}^{2}M_{\mathbf{H}}^{2})$ computational complexity of the lower bound. While Eq. \eqref{Differentinput:F} allows us to define different noise variances for each output and handle datasets observed at irregular input locations, it is also computationally more expensive to evaluate than Eq. \eqref{sameformula} as in practice we need to calculate the expectations $\Phi_{d}$, $\Psi_{d}$, $\psi_{d}$ for each output.

\section{Prediction}

In this section, we derive the predictive distribution of HMOGP-LV. 
For existing outputs and a test set of inputs $\mathbf{X}^{*}$, we have:

\begin{align}
q\left(\mathbf{f}^{*} \mid \mathbf{X}^{*}\right)=\int q\left(\mathbf{f}^{*} \mid \mathbf{X}^{*}, \mathbf{H}\right) q(\mathbf{H}) \mathrm{d} \mathbf{H}.
\label{predictionF}
\end{align}

Recalling Eq. \eqref{sparsegp}, the variational distribution in the integral can be analytically derived as:

\begin{align} 
q\left(\mathbf{f}^{*} \mid \mathbf{X}^{*}, \mathbf{H}\right) &=\int p\left(\mathbf{f}^{*} \mid \mathbf{U}, \mathbf{X}^{*}, \mathbf{H}\right) q\left(\mathbf{U}_{:}\right) \mathrm{d} \mathbf{U}_{:} =\mathcal{N}\left(\mathbf{f}^{*} \mid \tilde{\mathbf{m}}_*, \tilde{\mathbf{K}}_* \right), 
\end{align}

where $\tilde{\mathbf{m}}_*$ is $\mathbf{K}_{\mathbf{f}^{*} \mathbf{U}} \mathbf{K}_{\mathbf{U} \mathbf{U}}^{-1} \mathbf{M}_{:}$ and $\tilde{\mathbf{K}}_*$ is equal to $\mathbf{K}_{\mathbf{f}^{*} \mathbf{f}^{*}}-\mathbf{K}_{\mathbf{f}^{*} \mathbf{U}} \mathbf{K}_{\mathbf{U} \mathbf{U}}^{-1} \mathbf{K}_{\mathbf{f}^{*} \mathbf{U}}^{\top}$+ $\mathbf{K}_{\mathbf{f}^{*} \mathbf{U}}$ $\mathbf{K}_{\mathbf{U} \mathbf{U}}^{-1} \bm{\Sigma}^{\mathbf{U}_{:}} \mathbf{K}_{\mathbf{U} \mathbf{U}}^{-1} \mathbf{K}_{\mathbf{f}^{*} 
\mathbf{U}}^{\top}$
with $\mathbf{K}_{\mathbf{f}^{*} \mathbf{f}^{*}}=\mathbf{K}_{\mathbf{f}^{*} \mathbf{f}^{*}}^{H} \otimes \mathbf{K}_{\mathbf{f}^{*} \mathbf{f}^{*}}^{X}$ and $\mathbf{K}_{\mathbf{f}^{*} \mathbf{U}}=\mathbf{K}_{\mathbf{f}^{*} \mathbf{U}}^{H}  \otimes \mathbf{K}_{\mathbf{f}^{*} \mathbf{U}}^{X}$.
Although Eq. \eqref{predictionF} is intractable, we are still able to obtain the first and second moments of $\mathbf{f}^{*}$ in $q\left(\mathbf{f}^{*} \mid \mathbf{X}^{*}\right)$ \citep{titsias2010bayesian}.

\section{Experiments}

In this section, we evaluate HMOGP-LV on both synthetic and real-world datasets and compare its performance against alternative methods. 
The evaluation between competing approaches is performed regarding two performance metrics for regression problems: normalised mean square error (NMSE) and negative log predictive density (NLPD). Both for NMSE and NLPD, the smaller the values, the better.

\paragraph{Baselines:} In terms of structure assumptions, we compare our method with three GP models involving hierarchical kernel matrices as introduced in \cite{hensman2013hierarchical}, namely, \textbf{HGP} the original approach, \textbf{HGPInd} a modified version using inducing variables, and \textbf{DHGP} that presents a deep hierarchical structure.
Two multi-output GPs approaches are also considered: a standard linear model of coregionalisation (\textbf{LMC}) \citep{goovaerts1997geostatistics}, and the latent variables multi-output GPs model  (\textbf{LVMOGP}) \citep{dai2017efficient}. We also compared our method to a Neural Network (NN), with 2 layers of 200 units and a ReLU activation, to handle a single output. Both \textbf{HGP} and \textbf{HGPInd} can only handle a single output with its own replicas. \textbf{DHGP}, however, is able to deal with multiple outputs having their own replicas. \textbf{LMC} and \textbf{LVMOGP} can manage multiple outputs, but to deal with the multiple replicas per output, we stack them in concatenated vectors per output. The Adam optimiser \citep{kingma2014adam} is used for maximising the lower bound of the log marginal likelihood (i.e., $\mathcal{L}$ in Eq. \eqref{elbo}) with a 0.01 learning rate over 10,000 iterations.
The Adam optimiser is also used with identical settings to train \textbf{LMC} and \textbf{NN}.
The other models have been trained thanks to the L-BFGS-B algorithm implemented in SciPy \citep{virtanen2020scipy} over 10,000 iterations as well. We assume that each output has its own noise variance for all the models.
\begin{figure}
	\centering
	\includegraphics[width=0.65\textwidth]{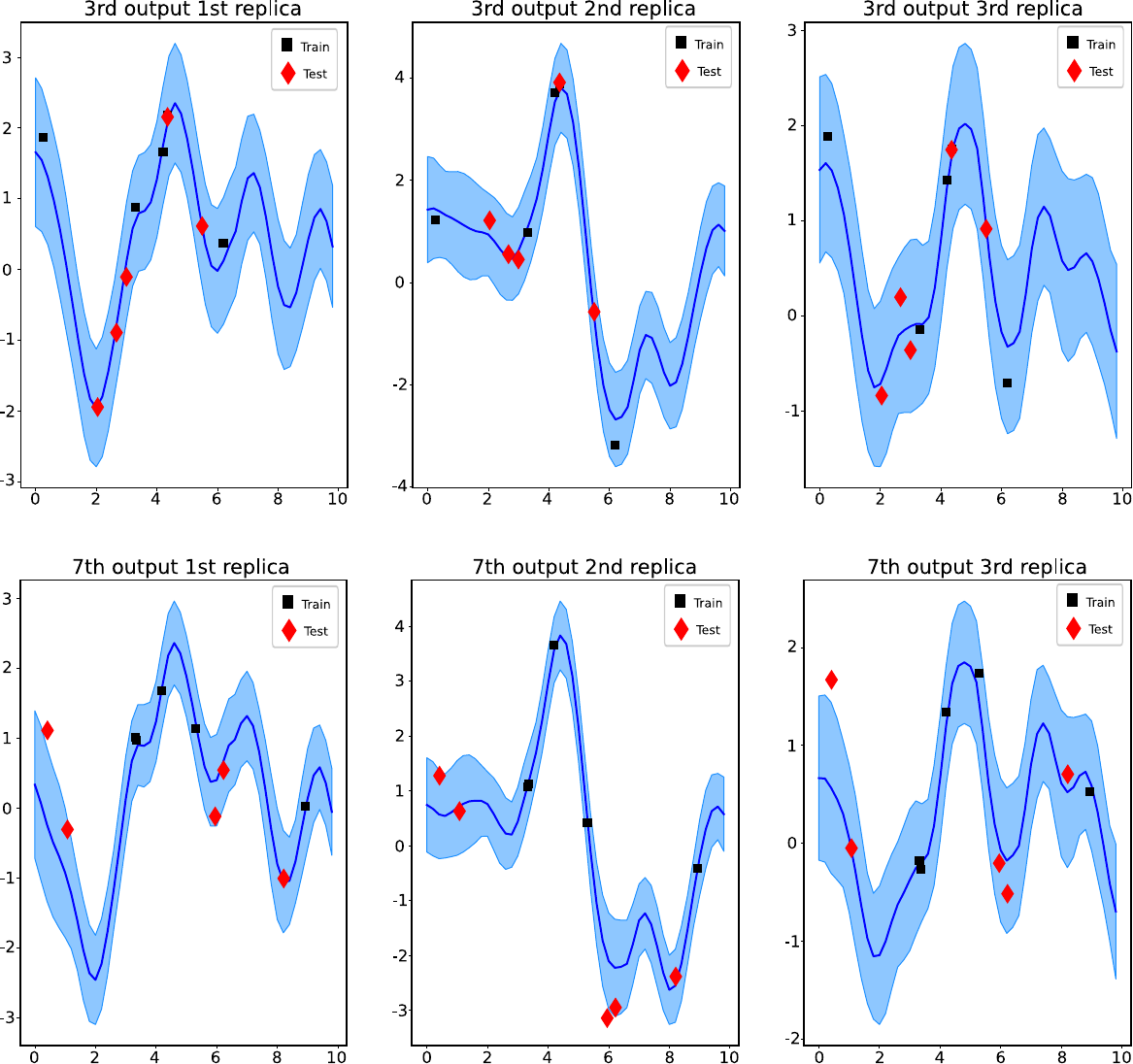}
	\caption{Mean predictive curves associated with their 95\% credible intervals for the third output (top row) and seventh output (bottom row) with three replicas each, coming from the synthetic dataset. Locations of training points (in black) and testing points (in red) are specific to each output.}
	\label{Plot:training_test}
\end{figure}

\paragraph{Computational Complexity:} Let us provide a quick discussion about the computational complexity of those different frameworks.
For the sake of simplicity, we assume here that all outputs are observed over the same input set, so the total number of data points is $N \times R$. Since \textbf{HMOGP-LV} is derived from \textbf{LVMOGP} with no extra computational burden, both methods present the same complexity, specifically, $\mathcal{O}\Big(\text{max}\left(NR,M_{\mathbf{H}}\right)\text{max}\left(D,M_{\mathbf{X}}\right)$ $\text{max}\left(M_{\mathbf{H}},M_{\mathbf{X}}\right)\Big)$ \citep{dai2017efficient}. 
Regarding \textbf{LMC}, the computational complexity is $\mathcal{O}\left(Q M^{3}+D NR Q M^{2}\right)$. 
The complexity of \textbf{HGP} and \textbf{HGPInd} is $\mathcal{O}\left((NR)^{3}\right)$ and $\mathcal{O}\left(NR(M_{\mathbf{H}}M_{\mathbf{X}})^{2}\right)$, respectively, whereas \textbf{DHGP} can generally be computed in $\mathcal{O}\left((DNR)^{3}\right)$ or reduced to $\mathcal{O}\left((ND)^{3}\right)$ in specific cases (see \cite{hensman2013hierarchical} for details).

All experiments were performed on a Dell PowerEdge C6320 with an Intel Xeon E5-2630 v3 at 2.40 GHz and 64GB of RAM\footnote{Our code is publicly available in the repository \url{https://github.com/ChunchaoPeter/HMOGP-LV}.}. Each experiment is repeated three times. Regarding the experiments with no missing replica, 50\% of the data points are dedicated to training in each replica and the other 50\% are used for testing purposes.
Neither \textbf{HGP} nor \textbf{DHGP} make use of inducing variables.
The value of $Q_{H}$ is set to 2 for \textbf{HMOGP-LV} and \textbf{LVMOGP} in all experiments.

\subsection{Simulation Study: Predicting Missing Time Points}

\begin{figure}
  \centering
 \includegraphics[width=0.8\textwidth]{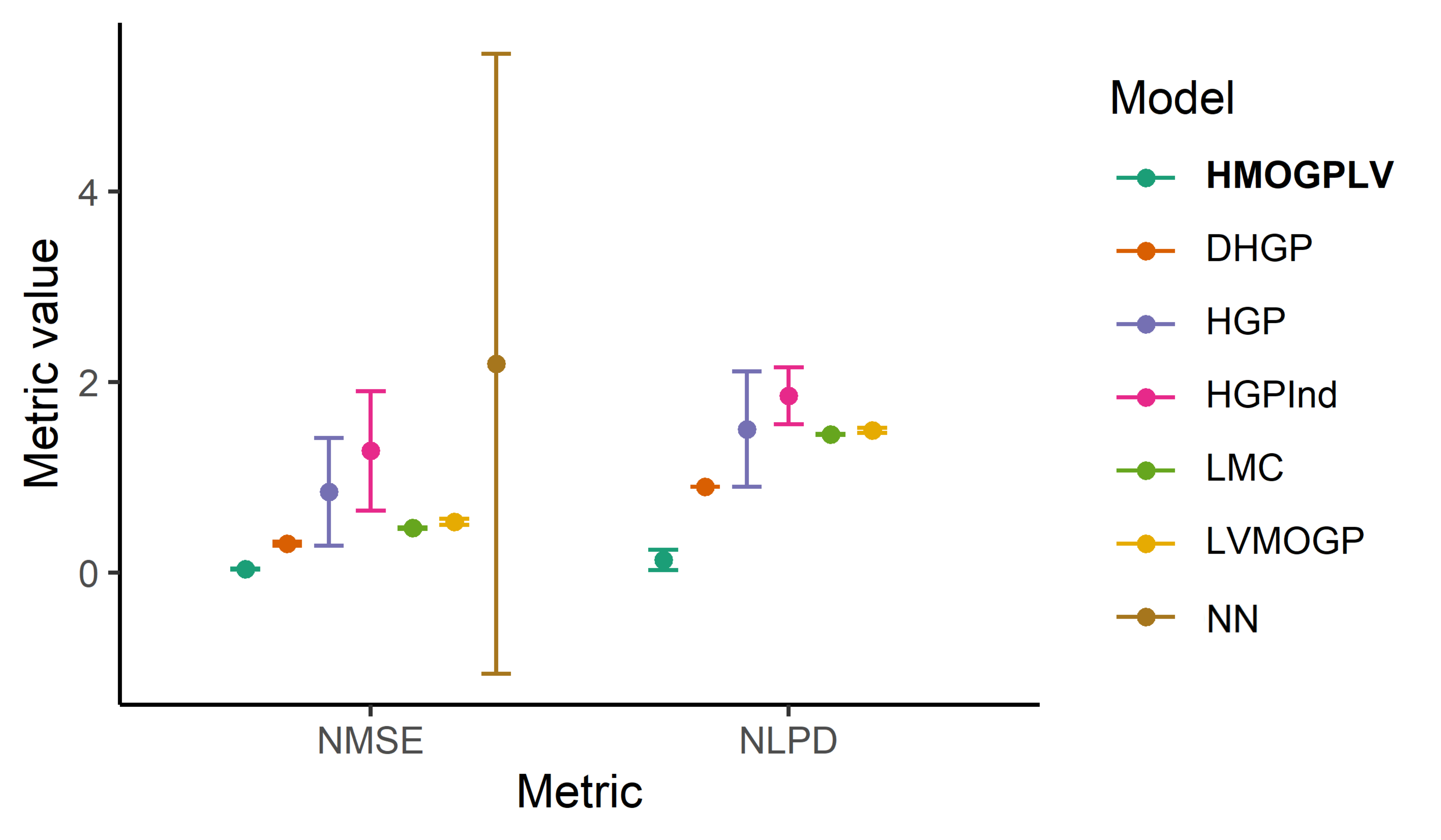}
	\caption{Prediction performances (mean $\pm$ standard deviation) for the first synthetic dataset. For both NMSE and NLPD values, the lower the better.}
	\label{Synthetic-Training-test}
\end{figure}
To exhibit the ability of our model to exploit correlations from hierarchical structures and between outputs simultaneously, we generated synthetic datasets by sampling from a Gaussian process with zero mean and covariance as in Eq. \eqref{HMOGPLV: Kernel_HMOGPLV}. 
This covariance function is a combination of two kernels: $k_H(\cdot,\cdot)$ for outputs (two-dimensional space) Kronecker-times a hierarchical kernel. 
Two kernels are also involved in the hierarchical kernel design: $k_g(\cdot,\cdot)$, which is assumed to be Mat\'{e}rn(3/2) with 1.0 lengthscale and 0.1 variance; and $k_f(\cdot,\cdot)$ defined as another Mat\'{e}rn(3/2) kernel with 1.0 lengthscale and 1.0 variance.
Each output is generated from a specific input set.
In addition, a Gaussian noise term with a 0.02 variance is added to each data sample. 
One synthetic dataset consists of 50 outputs with three replicas each, while each replica comprises 10 data points.

As an illustrative example, we displayed in Figure \ref{Plot:training_test} the prediction results for each replica in the third output (top row) and the seventh output (bottom row). One can notice in Figure \ref{Plot:training_test} that \textbf{HMOGP-LV} can offer remarkable predictions even from a handful of training points.
Our method provides both a mean prediction that closely fits testing points and an accurate uncertainty quantification encompassing relatively narrow regions around this curve. 
This desirable behaviour can be explained by the ability of  \textbf{HMOGP-LV} to share information at different levels by leveraging intra- and inter-output correlations and capturing the adequate hierarchical structure present in the data.
Sharing knowledge across different outputs allows for accurate predictions on unobserved regions for a specific replica while maintaining a relatively high level of confidence over all the input space considering such a sparse setting.
To pursue this simulation study, we provide in Figure \ref{Synthetic-Training-test} a comparative evaluation of predictive performances for all competing methods. 
It should be noticed that \textbf{HMOGP-LV} outperforms both single-output GP models (\textbf{HGP}, \textbf{HGPInd}), \textbf{NN} and multi-output ones (\textbf{LMC}, \textbf{LVMOGP}) in terms of NMSE and NLPD.

\begin{figure}
	\centering
	\includegraphics[width=0.9\textwidth]{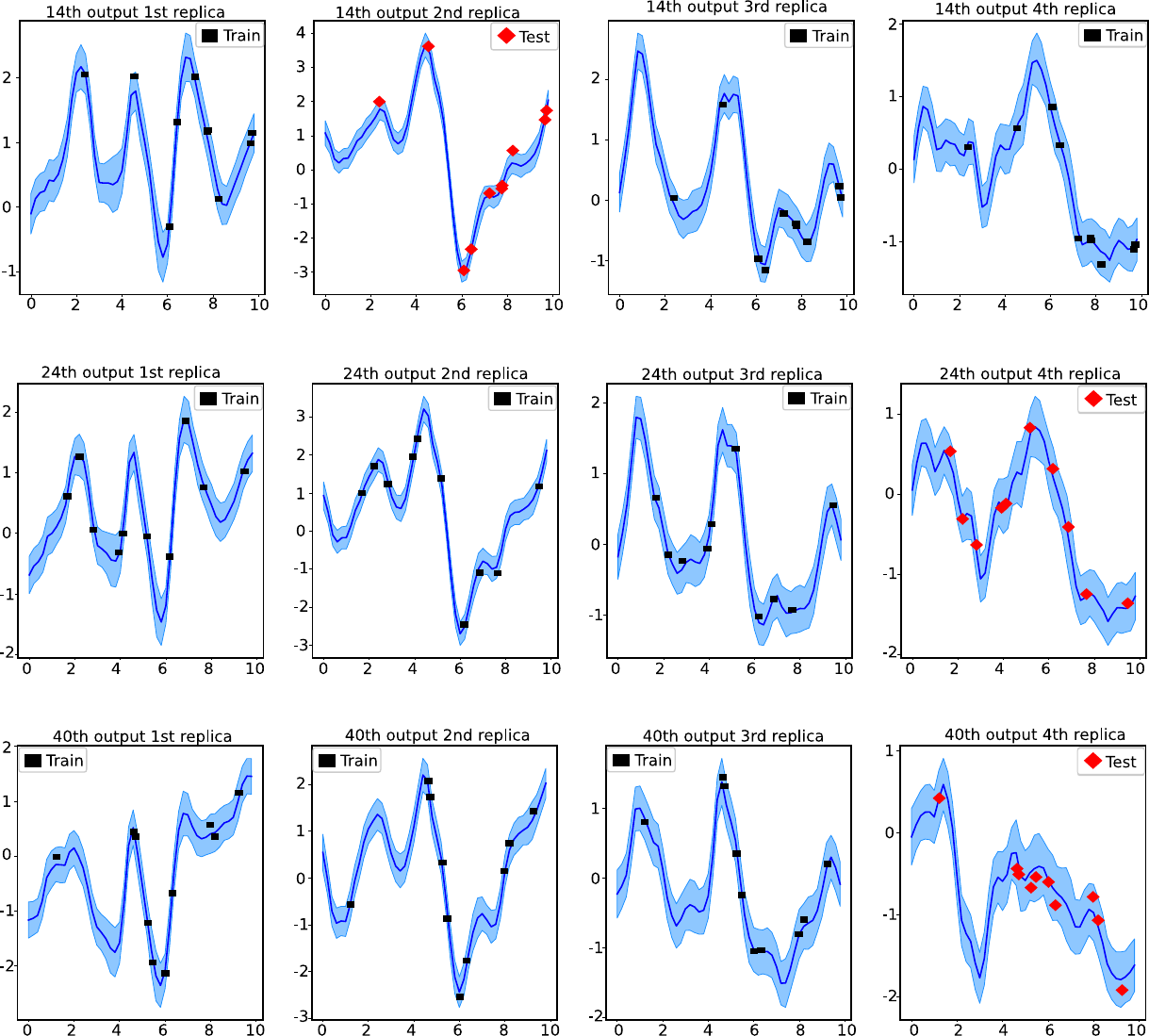}
	\caption{Top row: the result of the $14^{\text{th}}$ output with four replicas; Middle row: the result of the $24^{\text{th}}$ output with four replicas; Bottom row: the result of the $40^{\text{th}}$ output with four replicas. The black and red colour represents the train and test data points, respectively.}
	\label{Missing_synthetic_data}
\end{figure}

The best-performing method among the alternatives is \textbf{DHGP} as its deeper structure may approach the ability of our model to capture complex relationships at different levels. 
In particular, the top layer can capture correlations between different outputs, while the remaining two layers are likely to capture correlations among replicas.
Neither \textbf{LVMOGP} nor \textbf{LMC} offer satisfying results since they rely on a flat structure, preventing them from capturing the hierarchical structure of the dataset.
In the meantime, single-output GP methods remain limited as they cannot take advantage of other outputs to boost performances. Regarding \textbf{NN}, it presents lower performances in terms of RMSE and noticeably high variability in results. Moreover, \textbf{NN} does not provide uncertainty quantification and cannot be evaluated in terms of NLPD.
The ability of \textbf{HMOGP-LV} to exploit both properties simultaneously makes our model a sensible choice to handle this kind of highly nested dataset.

\subsection{Simulation Study: Predicting an Entirely Missing Replica}

\begin{figure}
	\centering
	\includegraphics[width=0.8\textwidth]{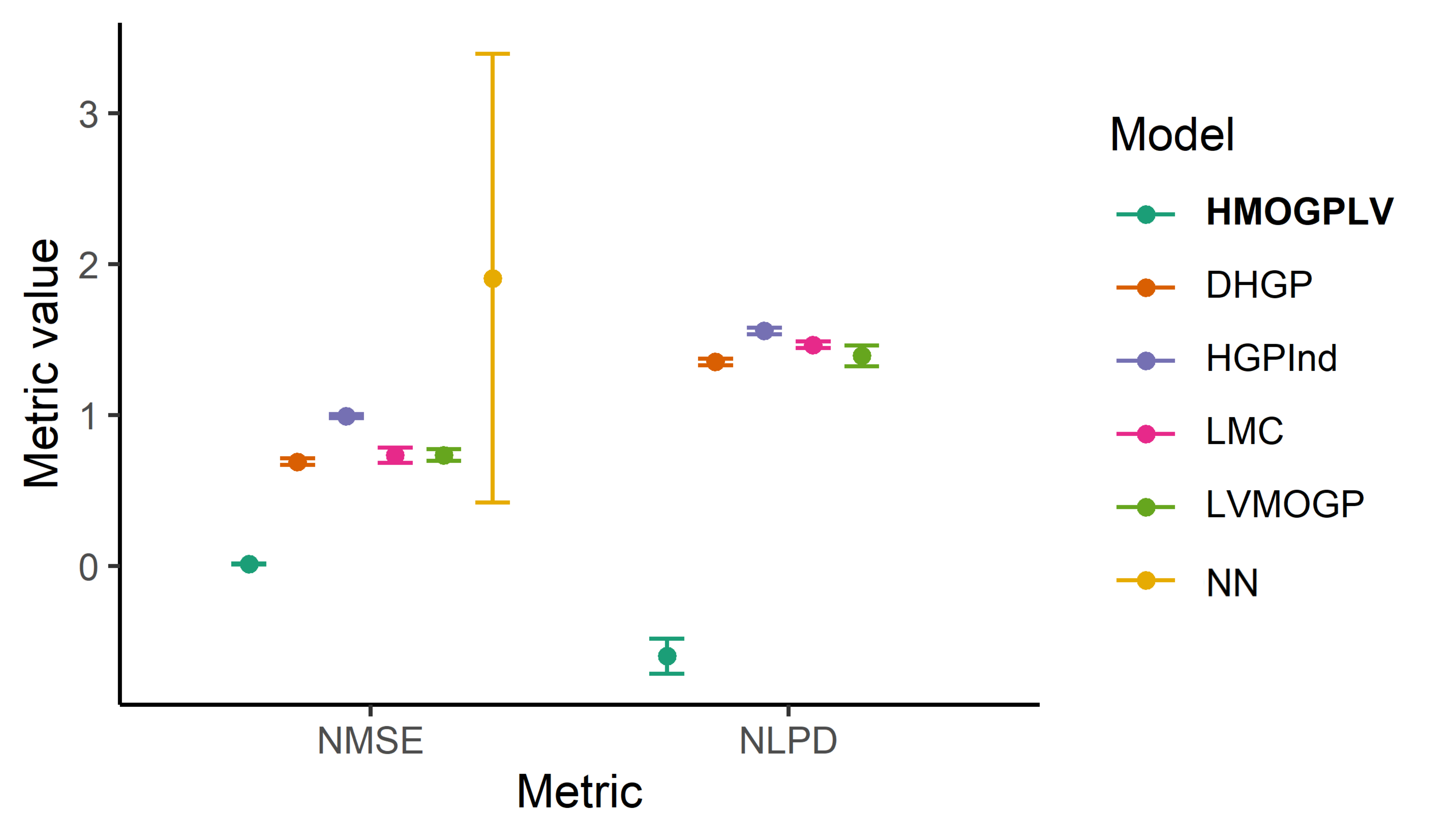}
	\caption{Prediction performances (mean $\pm$ standard deviation) for the second synthetic data with one missing replica in each output. For both NMSE and NLPD values, the lower the better.}
	\label{Synthetic-Missing-replicate}
\end{figure}
To demonstrate the unique ability of \textbf{HMOGP-LV} to predict an entirely missing replica, an additional experiment is provided with the following setting.
We generate 50 outputs with four replicas each, where each replica contains 10 data points.
In each output, we assume that one replica is missing.
Therefore, three replicas are used for training, and the remaining one is kept aside for testing purposes.

As an illustration, we display in Figure \ref{Missing_synthetic_data} the \textbf{HMOGP-LV} prediction for the three different outputs, where training points are in black and testing points in red.
For instance, with the $14^{\text{th}}$ output (top row), the first, third and fourth replicas are observed, whereas the second replica is missing.
One can observe in each case the excellent predictions for the missing replica.
In this example, this can probably be explained by the strong correlations among replicas in all outputs.
However, it confirms that our model adequately captures correlations and can transfer them through the inducing variables to predict the missing replica accurately.
In Figure \ref{Synthetic-Missing-replicate}, we compare our model against competitors for both evaluation metrics. Once again, \textbf{HMOGP-LV} offers superior performances compared to alternatives. Let us note that
\textbf{HGP} cannot make predictions for missing replicas as it is not originally designed to be trained in such settings. 
\textbf{HGPInd} uses other replicas in the same output to obtain the information for the missing replica and all the information kept in the inducing points. 
However, it cannot share knowledge across outputs whereas our model can fully leverage this information. 
Both \textbf{LVMOGP} and \textbf{LMC} can predict missing replicas since they do not distinguish replicas in each output that have a hierarchical structure.
Nevertheless, \textbf{HMOGP-LV} can keep information from all replicas in inducing variables to improve predictive performances.

\subsection{Real Datasets}

In this subsection, we compare the performance of \textbf{HMOGP-LV} against other GP models and \textbf{NN} on two real datasets, related to genomics and motion capture applications for multi-output regression problems.

\subsubsection{Gene Dataset}

The first problem we aim at tackling consists in predicting temporal gene expression of Drosophila development based on a dataset originally proposed by \cite{kalinka2010gene}. For each of the six observed Drosophila species, the expression of 3695 genes has been measured in eight replicas at different time points.
Following \cite{hensman2013hierarchical}, this paper focuses on one of these six species (\emph{melanogaster}) and the following genes considered as outputs in our model: `CG12723', `CG13196', `CG13627', `Osi15'.
For those outputs, each of the eight replicas is partially observed on a grid of 10 distinct time points (i.e. each replica has a specific set of inputs, which is a sub-sample of a 10-point common grid).
\begin{figure}
	\centering
	\includegraphics[width=\textwidth]{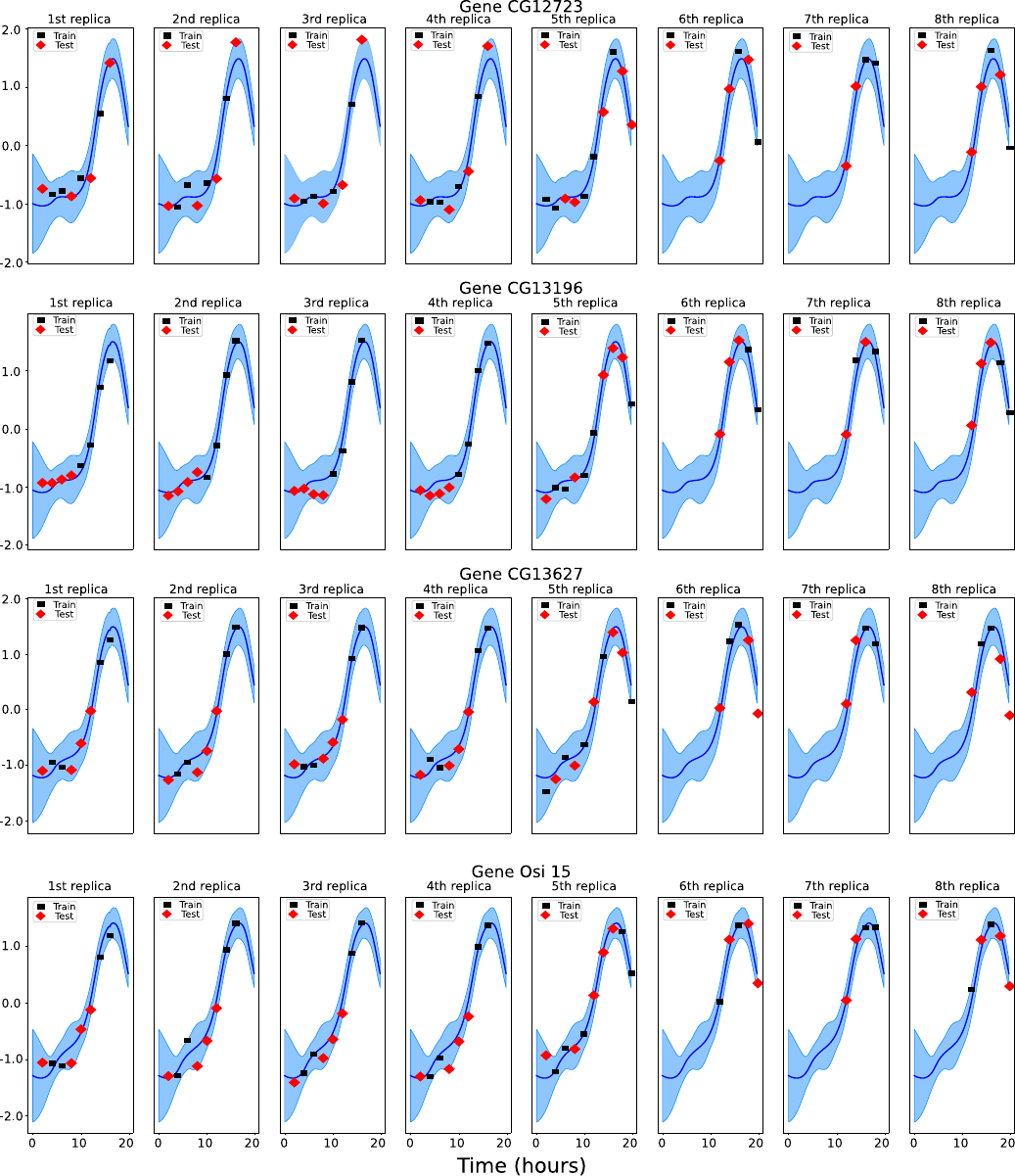}
	\caption{Mean predictive curves associated with their 95\% credible intervals for all outputs and replicas of the gene dataset. Locations of training points (in black) and testing points (in red) are specific to each output.}
	\label{Gene-prediction}
\end{figure}
When considering such relatively small datasets, setting the value of $M_{\mathbf{X}}$ to 14 for \textbf{HMOGP-LV}, \textbf{HGPInd}, \textbf{LVMOGP}, and \textbf{LMC} appeared as a sensible choice.
In the case of \textbf{HMOGP-LV} and \textbf{LVMOGP}, we additionally defined $M_{\mathbf{H}} = 2$. 
As previously mentioned, the goal of this experiment consists in predicting 50\% of the data points that have been randomly removed in each replica to be used as testing points. 
To illustrate the behaviour of our method to tackle such a task, we display in Figure \ref{Gene-prediction} the GP predictions obtained by applying \textbf{HMOGP-LV} on all outputs and replicas.
It can be noticed that in all cases, the mean curve sticks close to the true test points while maintaining narrow credible intervals on the studied domain, though uncertainty significantly increases when moving towards 0 as the number of observed data is low for all replicas.
\begin{figure}
	\centering
	\includegraphics[width=0.8\textwidth]{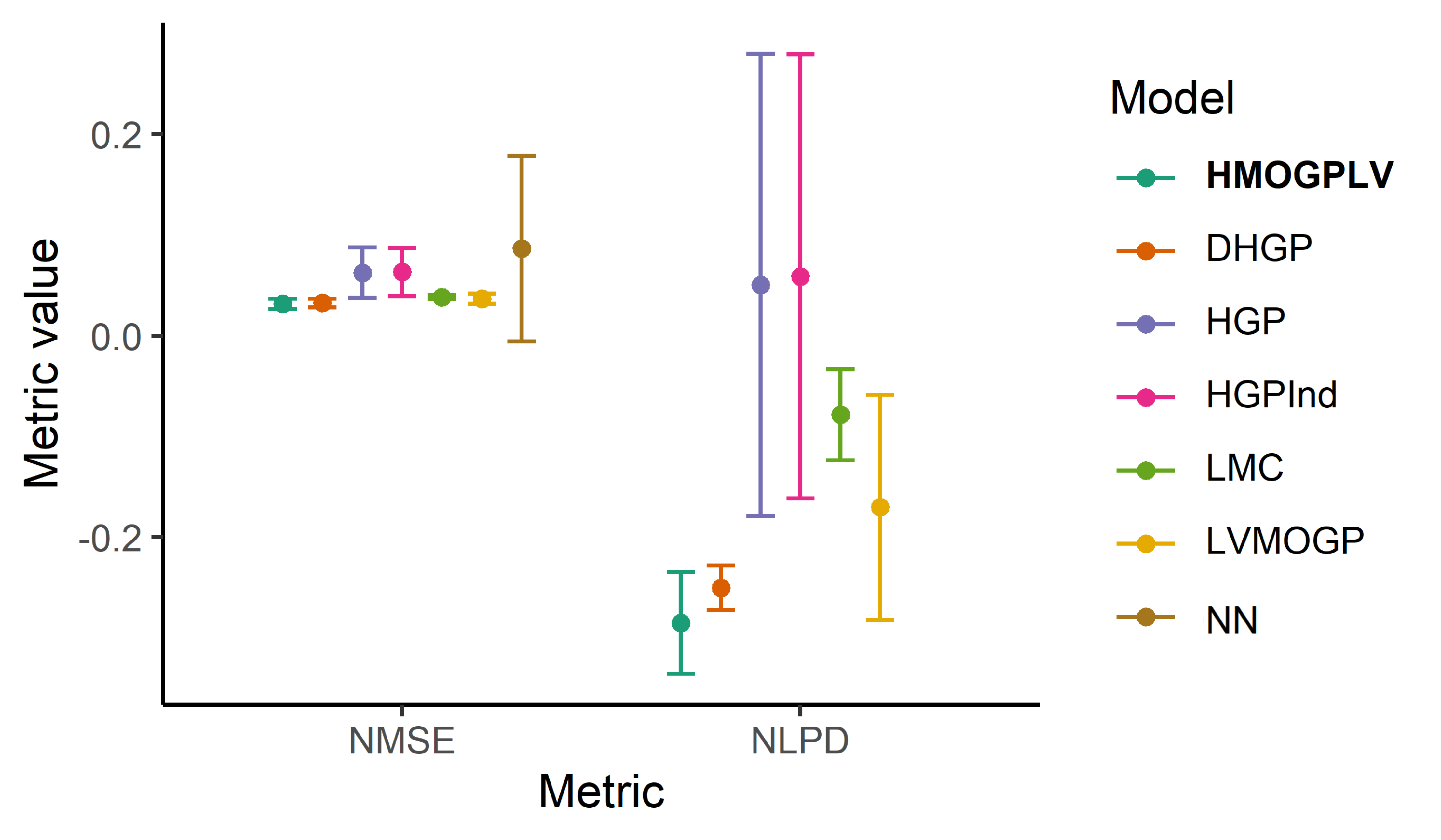}
	\caption{Prediction performances (mean $\pm$ standard deviation)  for the gene dataset. For both NMSE and NLPD values, the lower the better.}
	\label{Gene-Training-test}
\end{figure}
While this visual inspection is promising, the comparison with competing methods provided in Figure \ref{Gene-Training-test} highlights that \textbf{HMOGP-LV} also outperforms the alternatives. 
Let us mention that \textbf{DHGP} offers once again performances that are noticeably better than other approaches, confirming our first insights from the synthetic data experiments.

As previously mentioned during modelling developments, our method also allows the prediction of an entirely missing replica, by sharing information across outputs and replicas to reconstruct the signal. 
We propose this additional experiment applied to the gene dataset in supplementary materials, and demonstrate the remarkable ability of \textbf{HMOGP-LV} to provide predictions that remain accurate even in the absence of data points for a whole replica. 

\subsubsection{Motion Capture Database}

\begin{figure}
	\centering
	\includegraphics[width=0.75\textwidth]{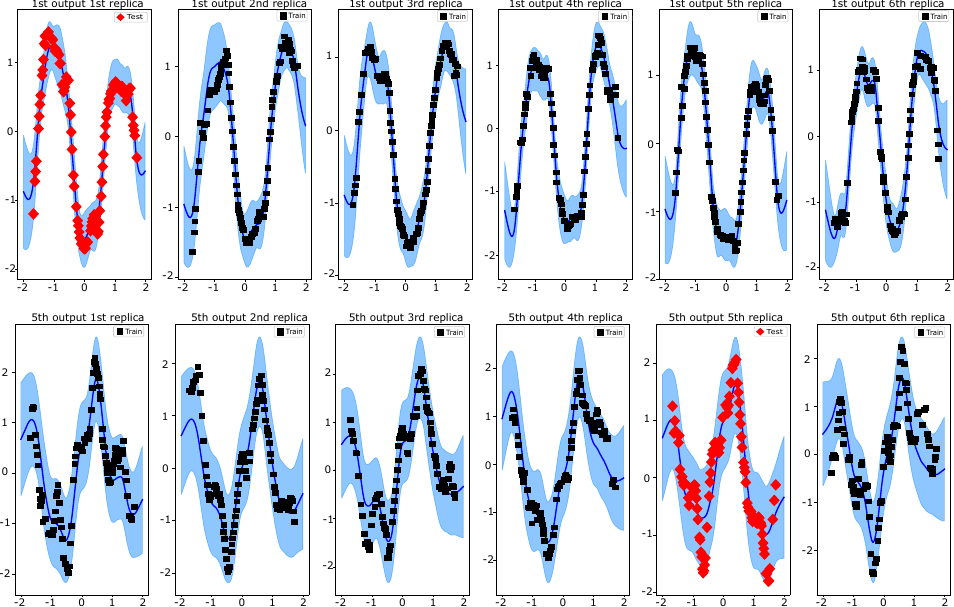}
	\caption{Mean predictive curves associated with their 95\% credible intervals for all outputs and replicas of the MOCAP-9 dataset. Locations of training points (in black) and testing points (in red) are specific to each output.}
	\label{MOCAP-Missing-prediction}
\end{figure}

Let us pursue by presenting another application of \textbf{HMOGP-LV} involving observations from the CMU motion capture database (MOCAP)
\footnote{The CMU Graphics Lab Motion Capture Database was created with funding from NSF EIA-0196217 and is available at \url{http://mocap.cs.cmu.edu}.}.
In this dataset, four different categories of movement are identified and distinguished: walking, running, golf swing and jumping. 
According to the experimental setting, only specific parts of the body are tracked by the motion capture devices.
Regarding walking, the data of interest consists of trials number 2, 3, 8 and 9, for the $8$-th subject, where we consider each trial as a replica.
Our study focuses on right-hand movements (humerus, radius wrist, femur and tibia) for which we consider 16 positions in total.
Additionally, the input and output data points are both scaled to have a zero mean and unit variance.
Each position is identified as an output, though we only retained outputs with a signal-to-noise ratio over 20 dB. 
Therefore, using 16 outputs, each of them containing four replicas, and designated as \emph{MOCAP-8}. 
For the case of running, data for the $9$-th subject were extracted for trials number 1, 2, 3, 5, 6, and 11.
Head and foot movements (lower-neck, upper-neck, head, femur, tibia and foot) were tracked, for a total of 16 outputs with six replicas each (\emph{MOCAP-9}). 
The golfswing case is studied through trials number 3, 4, 5, 7, 8 and 9 of the $64$-th individual. 
We consider left and right-hand movements (humerus, radius and wrist) by modelling nine outputs with six replicas each (\emph{MOCAP-64}).
Finally, jumping is analysed through trials number 3, 4, 11 and 17 of the $118$-th individual. 
We chose to focus of foot movements (femur, tibia and foot) to collect 12 outputs with four replicas each (\emph{MOCAP-118}). 
The overall parameter settings are summarised within a table in supplementary materials. 
In all settings, each replica is observed over 200 time points except MOCAP-9 (in MOCAP-9, each replica is observed over 100 time points since its replica has around 140 times).
In this experiment, we aim to predict unobserved replicas.
More precisely, for each output, one of its replicas is entirely missing, while all the others are fully observed. 
\begin{figure}[ht]
	\centering
	\includegraphics[width=0.75\textwidth]{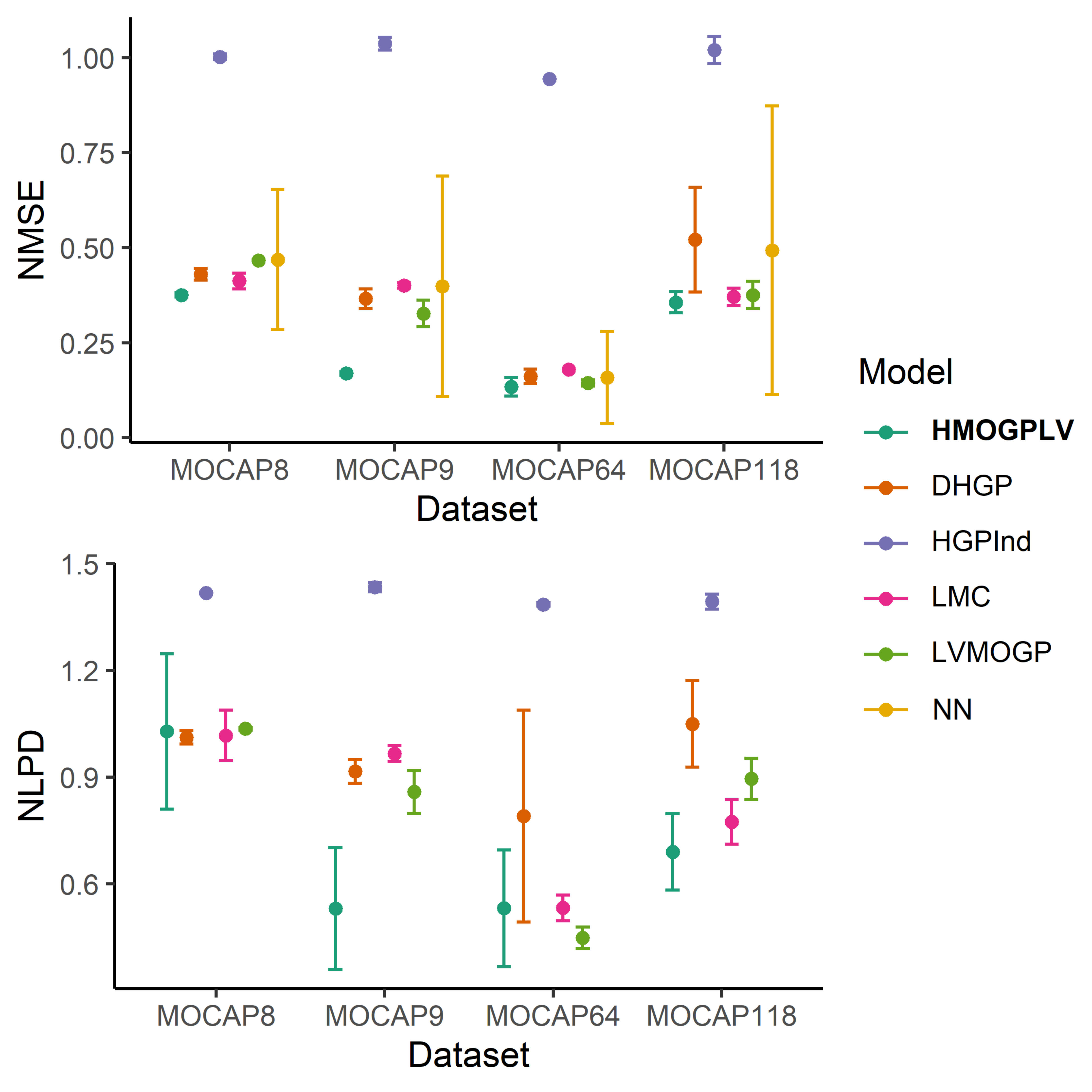}
	\caption{Prediction performances (mean $\pm$ standard deviation) for the MOCAP-8, MOCAP-9, MOCAP-64 and MOCAP-118 datasets. For both NMSE and NLPD values, the lower the better.}
	\label{MOCAP-Missing-replicate}
\end{figure}
As highlighted in Figure \ref{MOCAP-Missing-replicate}, \textbf{HMOGP-LV} outperforms other methods in most situations, except for MOCAP-64 and MOCAP-118 in terms of NMSE, where \textbf{DHGP} and \textbf{LVMOGP} present comparable results. 
In particular, the results of the MOCAP-9 experiment, for which the improvement provided by our method is the most prominent, are illustrated in Figure \ref{MOCAP-Missing-prediction}. 
One can notice how our model retrieves adequately the overall pattern for the missing replica at no cost in terms of uncertainty. 
As displayed, it seems that sharing information at different levels, both among outputs and replicas, allows the prediction to remain accurate regardless of the sub-sample of data that is removed.
It is worth mentioning that both multi-output methods (\textbf{LMC} and \textbf{LVMOGP}) also exhibit excellent performance in those task, although \textbf{HMOGP-LV} seems to remain the most sensible choice overall.

\section{Conclusion}
In this paper, we introduced HMOGP-LV, an extended framework of multi-output Gaussian processes to deal with multiple regression problems for hierarchically structured datasets. HMOGP-LV uses latent variables to capture the correlation between multiple outputs and a hierarchical kernel matrix to capture the dependency between replicas for each output. Even in the presence of missing replicas, HMOGP-LV remains able to make predictions by using information shared through inducing variables. We experimentally demonstrated that HMOGP-LV offers enhanced performances in terms of NMSE and NLPD compared to natural competitors for both synthetic and real datasets. 
In terms of limitations, HMOGP-LV only addresses regression problems so far since the likelihood considered is Gaussian. Moreover, our model is also limited to two layers of hierarchy when accounting for correlations. Therefore, several extensions of the present framework would be valuable, such as enabling heterogeneous multi-output prediction \citep{moreno2018heterogeneous} or defining additional layers to build a deeper hierarchical structure \citep{hensman2013hierarchical}.

\clearpage

\section*{CRediT authorship contribution statement}

\textbf{Chunchao Ma}: Methodology, Software, Writing – original draft. \textbf{Arthur Leroy}: Investigation, Formal analysis, Writing – review \& editing. \textbf{Mauricio \'{A}lvarez}: Conceptualization, Writing – review \& editing, Supervision.

\section*{Declaration of competing interest}

The authors declare that they have no known competing
financial interests or personal relationships that could have
appeared to influence the work reported in this paper.

\section*{Data availability}
The CMU Graphics Lab Motion Capture  (MOCAP) Database was created with funding from NSF EIA-0196217 and is available at \url{http://mocap.cs.cmu.edu}.
The gene dataset is available in this repository: \url{https://github.com/ChunchaoPeter/HMOGP-LV/tree/main/Gene_data_set}.

\section*{Code availability}
The Python implementation of \textbf{HMOGP-LV} is freely available in the following repository: \url{https://github.com/ChunchaoPeter/HMOGP-LV}.

\section*{Acknowledgements}

Chunchao Ma would like to thank Zhenwen Dai for the helpful conversations. Arthur Leroy and Mauricio \'{A}lvarez have been financed by the Wellcome Trust project 217068/Z/19/Z


\appendix

\section{Proofs}

\begin{figure}
	\centering
	\includegraphics[width=0.9\textwidth]{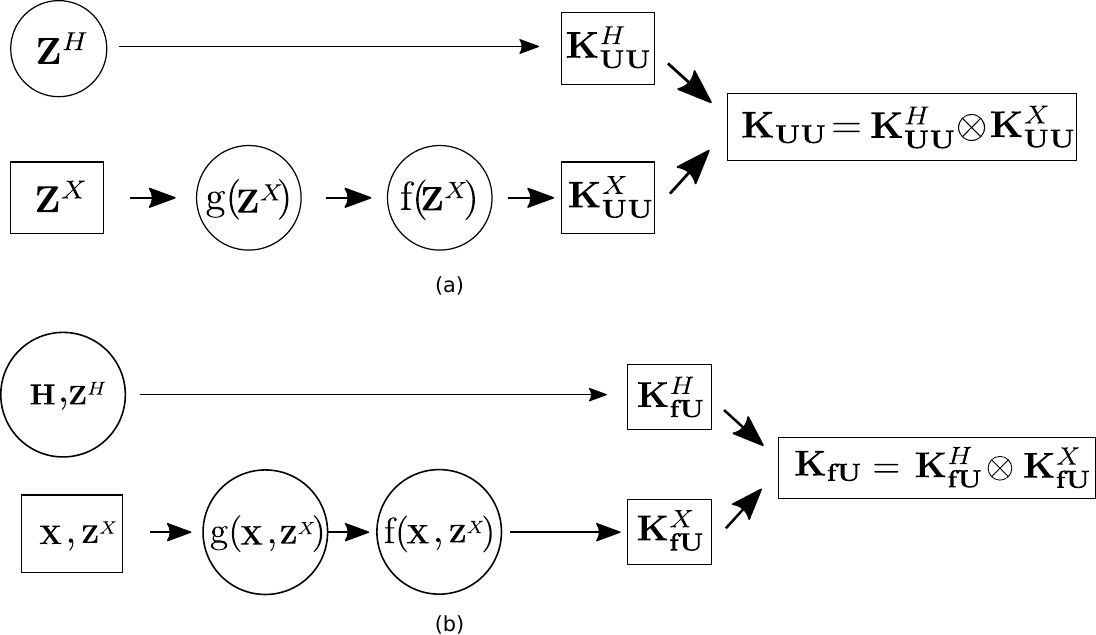}
	\caption{(a): Summary of the procedure used to derive the
  kernel matrix for inducing variables, where $\mathbf{Z}^{X}$ and $\mathbf{Z}^{H}$ are associated with the inputs  $\mathbf{X}$ and the latent variables $\mathbf{H}$, respectively; (b): Summary of the procedure used to derive the kernel matrix between observations and inducing variables.}
	\label{Kernel_final_inducing_variables}
\end{figure}

In this section, we present technical details for deriving the lower bound of the log marginal likelihood as well as computationally efficient formulations by exploiting Kronecker product decomposition for $\mathcal{F}$. Before diving into the mathematical details, let us also provide in Figure \ref{Kernel_final_inducing_variables} an illustrative recall of the modelling assumptions.

\subsection{Derivation of the Log-marginal Likelihood Lower Bound}
\label{appndx:HMOGPLV-ELBO}
To obtain the lower bound of the log marginal likelihood of our model, we assume that the variational posterior distributions are $q(\mathbf{H})$, $q(\mathbf{U}_{:})$ and $q(\mathbf{f} \mid \mathbf{U}_{:}, \mathbf{H})=p(\mathbf{f} \mid \mathbf{U}_{:}, \mathbf{H})$, such as:
\begin{align}
\text{log } p\left(\mathbf{y} \right) &= \text{log} \int \int \int p\left(\mathbf{y}, \mathbf{f}, \mathbf{H}, \mathbf{U}_{:}\right)\text{d}\mathbf{f}\text{d}\mathbf{H}\text{d}\mathbf{U}_{:} \nonumber \\
&=\text{log} \int \int \int \frac{p\left(\mathbf{y}, \mathbf{f}, \mathbf{H}, \mathbf{U}_{:} \right) q\left(\mathbf{f}, \mathbf{H}, \mathbf{U}_{:} \right)}{q\left(\mathbf{f}, \mathbf{H}, \mathbf{U}_{:} \right)}\text{d}\mathbf{f}\text{d}\mathbf{H}\text{d}\mathbf{U}_{:} \nonumber \\
&\geq \int \int \int q\left(\mathbf{f}, \mathbf{H}, \mathbf{U}_{:} \right) \text{log}  \frac{p\left(\mathbf{y}, \mathbf{f}, \mathbf{H}, \mathbf{U}_{:} \right)}{q\left(\mathbf{f}, \mathbf{H}, \mathbf{U}_{:} \right)}\text{d}\mathbf{f}\text{d}\mathbf{H}\text{d}\mathbf{U}_{:} \nonumber \\
& = \mathcal{L}.
\end{align}
\begin{align}
\mathcal{L}=&\left\langle\log \frac{p\left(\mathbf{y}, \mathbf{f}, \mathbf{H}, \mathbf{U}_{:}\right)}{q\left(\mathbf{f}, \mathbf{H}, \mathbf{U}_{:} \right)}\right\rangle_{q\left(\mathbf{f}, \mathbf{H}, \mathbf{U}_{:} \right)} \nonumber \\
=&\int \int \int p \left(\mathbf{f} \mid \mathbf{U}_{:}, \mathbf{H}\right) q(\mathbf{U}_{:}) q(\mathbf{H})  \nonumber \\
& \text{log} \frac{p(\mathbf{y}|\mathbf{f}, \mathbf{H},\mathbf{U}_{:})p \left(\mathbf{f} \mid \mathbf{U}_{:}, \mathbf{H}\right) p(\mathbf{U}_{:}) p(\mathbf{H})}{p \left(\mathbf{f} \mid \mathbf{U}_{:}, \mathbf{H}\right) q(\mathbf{U}_{:}) q(\mathbf{H})}\text{d} \mathbf{f}\text{d}\mathbf{U}_{:}\text{d}\mathbf{H} \nonumber \\
=&\int \int \int p \left(\mathbf{f} \mid \mathbf{U}_{:}, \mathbf{H}\right) q(\mathbf{U}_{:}) q(\mathbf{H}) \text{log} \frac{p(\mathbf{y}|\mathbf{f}, \mathbf{H},\mathbf{U}_{:}) p(\mathbf{U}_{:}) p(\mathbf{H})}{q(\mathbf{U}_{:}) q(\mathbf{H})}\text{d}\mathbf{f}\text{d}\mathbf{U}_{:}\text{d}\mathbf{H}.
\end{align}
Finally, 
\begin{align} \mathcal{L} &=\int q(\mathbf{H})\left[\int q(\mathbf{U}_{:})\left[\mathbb{E}_{p(\mathbf{f} \mid \mathbf{U}_{:}, \mathbf{H})}[\log p(\mathbf{y} \mid \mathbf{f}, \mathbf{H})]+\log \frac{p(\mathbf{U}_{:})}{q(\mathbf{U}_{:})}+\log \frac{p(\mathbf{H})}{q(\mathbf{H})}\right]\text{d}\mathbf{U}_{:}\right]  \nonumber \\
& \quad \quad \text{d} \mathbf{H} \nonumber \\
&=\overbrace{\mathbb{E}_{q(\mathbf{f}, \mathbf{U}_{:}, \mathbf{H})}[\log p(\mathbf{y} \mid \mathbf{f}, \mathbf{H})]}^{\mathcal{F}}-\mathrm{K L} (q(\mathbf{H}) \| p(\mathbf{H}))-\mathrm{KL}(q(\mathbf{U}_{:}) \| p(\mathbf{U}_{:})).
\end{align}

\subsection{Derivation of \texorpdfstring{$\mathcal{F}$}{\texttwoinferior} Given the Same Input Datasets}
\label{appndx:HMOGPLV-ExpectationFindsameoutput}

In this section, we show details for deriving $\mathcal{F}$ using the same input datasets:
\begin{align} \mathcal{F} &= \mathbb{E}_{p \left(\mathbf{f} \mid \mathbf{U}_{:}, \mathbf{H}\right) q(\mathbf{U}_{:}) q(\mathbf{H})}\left[\log p\left(\mathbf{y} \mid \mathbf{f}, \mathbf{H}\right)\right] \nonumber \\ 
&= \int q(\mathbf{H}) \int q\left(\mathbf{U}_{:}\right) \underbrace{\int p \left(\mathbf{f} \mid \mathbf{U}_{:}, \mathbf{H}\right) \log p\left(\mathbf{y} \mid \mathbf{f}, \mathbf{H}\right)\text{d}\mathbf{f}}_{\mathcal{L}_{F}}\text{d}\mathbf{U}_{:}\text{d}\mathbf{H} \nonumber \\ 
&= \int q(\mathbf{H}) \underbrace{\int q\left(\mathbf{U}_{:}\right) \mathcal{L}_{F}\text{d}\mathbf{U}_{:}}_{\mathcal{L}_{U}}\text{d}\mathbf{H} \nonumber \\
 &=\underbrace{\int q(\mathbf{H}) \mathcal{L}_{U}\text{d}\mathbf{H}}_{\mathcal{L}_{H}}. 
\end{align}
First, we calculate $\mathcal{L}_{F}$:
\begin{align} 
\mathcal{L}_{F} = & \int  p \left(\mathbf{f} \mid \mathbf{U}_{:}, \mathbf{H}\right) \log p\left(\mathbf{y} \mid \mathbf{f}, \mathbf{H} \right)\text{d}\mathbf{f} \nonumber \\ 
= & \log \mathcal{N}\left(\mathbf{y} \mid \mathbf{K}_{\mathbf{f} \mathbf{U}} \mathbf{K}_{\mathbf{U} \mathbf{U}}^{-1} \mathbf{U}_{:}, \sigma^{2}\right) - \frac{1}{2\sigma^{2}} \text{Tr} \left[\mathbf{K}_{\mathbf{f} \mathbf{f}}-\mathbf{K}_{\mathbf{f} \mathbf{U}} \mathbf{K}_{\mathbf{U} \mathbf{U}}^{-1} \mathbf{K}_{\mathbf{f} \mathbf{U}}^{\top} \right],
\end{align}
where $ p\left(\mathbf{f} \mid \mathbf{U}_{:}, \mathbf{H}\right)=\mathcal{N}\left(\mathbf{f} \mid \mathbf{K}_{\mathbf{f} \mathbf{U}} \mathbf{K}_{\mathbf{U} \mathbf{U}}^{-1} \mathbf{U}_{:}, \mathbf{K}_{\mathbf{f} \mathbf{f}}-\mathbf{K}_{\mathbf{f} \mathbf{U}} \mathbf{K}_{\mathbf{U} \mathbf{U}}^{-1} \mathbf{K}_{\mathbf{f} \mathbf{U}}^{\top}\right)$ and $\text{Tr}[\cdot]$ is a trace of a matrix. Second, we calculate $\mathcal{L}_{U}$:
\begin{align}
\mathcal{L}_{U} = & \int q\left(\mathbf{U}_{:}\right) \mathcal{L}_{F}\text{d}\mathbf{U}_{:} \nonumber \\
= & \log \mathcal{N}\left(\mathbf{y} \mid \mathbf{K}_{\mathbf{f} \mathbf{U}} \mathbf{K}_{\mathbf{U} \mathbf{U}}^{-1} \mathbf{M}_{:}, \sigma^{2}\right) - \frac{1}{2\sigma^{2}} \text{Tr} \left[\mathbf{K}_{\mathbf{f} \mathbf{f}}-\mathbf{K}_{\mathbf{f} \mathbf{U}} \mathbf{K}_{\mathbf{U} \mathbf{U}}^{-1} \mathbf{K}_{\mathbf{f} \mathbf{U}}^{\top}\right] \nonumber \\
& - \frac{1}{2\sigma^{2}} \text{Tr} \left[\bm{\Sigma}^{\mathbf{U}_{:}} \mathbf{K}_{\mathbf{U} \mathbf{U}}^{-1} \mathbf{K}_{\mathbf{f} \mathbf{U}}^{\top}\mathbf{K}_{\mathbf{f} \mathbf{U}}\mathbf{K}_{\mathbf{U} \mathbf{U}}^{-1}\right].
\end{align}
where $q(\mathbf{U}_{:})=\mathcal{N}\left(\mathbf{U}_{:} \mid \mathbf{M}_{:}, \bm{\Sigma}^{\mathbf{U}_{:}}\right)$ in which $\mathbf{U}_{:}$ and $\mathbf{M}_{:}$ are variational parameters. Finally, we consider $\mathcal{L}_{H}$:
\begin{align}
\mathcal{L}_{H}  = &\int q(\mathbf{H}) \mathcal{L}_{U}\text{d}\mathbf{H} \nonumber \\
= & \left\langle\log \mathcal{N}\left(\mathbf{y} \mid \mathbf{K}_{\mathbf{f} \mathbf{U}} \mathbf{K}_{\mathbf{U} \mathbf{U}}^{-1} \mathbf{M}_{:}, \sigma^{2}\right)\right\rangle_{q(\mathbf{H})}  \nonumber \\
& - \frac{1}{2\sigma^{2}} \text{Tr} \left[\left\langle\mathbf{K}_{\mathbf{f} \mathbf{f}}\right\rangle_{q(\mathbf{H})}- \mathbf{K}_{\mathbf{U} \mathbf{U}}^{-1} \left\langle\mathbf{K}_{\mathbf{f} \mathbf{U}}^{\top}\mathbf{K}_{\mathbf{f} \mathbf{U}}\right\rangle_{q(\mathbf{H})}\right] \nonumber \\
&- \frac{1}{2\sigma^{2}} \text{Tr} \left[\bm{\Sigma}^{\mathbf{U}_{:}} \mathbf{K}_{\mathbf{U} \mathbf{U}}^{-1} \left\langle\mathbf{K}_{\mathbf{f} \mathbf{U}}^{\top}\mathbf{K}_{\mathbf{f} \mathbf{U}}\right\rangle_{q(\mathbf{H})}\mathbf{K}_{\mathbf{U} \mathbf{U}}^{-1}\right] \nonumber \\
= & \underbrace{-\frac{DNR}{2} \log 2 \pi \sigma^{2}-\frac{1}{2 \sigma^{2}} \mathbf{y}^{\top} \mathbf{y}}_{\textbf{C}} + \frac{1}{\sigma^{2}} \mathbf{y}^{\top} \underbrace{ \left\langle\mathbf{K}_{\mathbf{f} \mathbf{U}}\right\rangle_{q(\mathbf{H})}}_{\Psi}\mathbf{K}_{\mathbf{U} \mathbf{U}}^{-1} \mathbf{M}_{:} \nonumber \\
& - \frac{1}{2 \sigma^{2}} \mathbf{M}_{:}^{\top}\mathbf{K}_{\mathbf{U} \mathbf{U}}^{-1} \underbrace{\left\langle\mathbf{K}_{\mathbf{f} \mathbf{U}}^{\top}\mathbf{K}_{\mathbf{f} \mathbf{U}}\right\rangle_{q(\mathbf{H})}}_{\Phi}\mathbf{K}_{\mathbf{U} \mathbf{U}}^{-1}\mathbf{M}_{:} \nonumber - \frac{1}{2 \sigma^{2}} \underbrace{\text{Tr} \left\langle\mathbf{K}_{\mathbf{f} \mathbf{f}}\right\rangle_{q(\mathbf{H})}}_{\psi} \nonumber \\ 
& + \frac{1}{2\sigma^{2}} \text{Tr}\left[\mathbf{K}_{\mathbf{U} \mathbf{U}}^{-1} \underbrace{\left\langle\mathbf{K}_{\mathbf{f} \mathbf{U}}^{\top}\mathbf{K}_{\mathbf{f} \mathbf{U}}\right\rangle_{q(\mathbf{H})}}_{\Phi}\right]  \nonumber \\
& - \frac{1}{2\sigma_{d}^{2}} \text{Tr} \left[\bm{\Sigma}^{\mathbf{U}_{:}} \mathbf{K}_{\mathbf{U} \mathbf{U}}^{-1} \underbrace{\left\langle\mathbf{K}_{\mathbf{f} \mathbf{U}}^{\top}\mathbf{K}_{\mathbf{f} \mathbf{U}}\right\rangle_{q(\mathbf{H})}}_{\Phi}\mathbf{K}_{\mathbf{U} \mathbf{U}}^{-1}\right] \nonumber \\
= & \textbf{C} + \frac{1}{\sigma^{2}} \mathbf{y}^{\top} \Psi \mathbf{K}_{\mathbf{U} \mathbf{U}}^{-1} \mathbf{M}_{:} - \frac{1}{2 \sigma^{2}} \left( \psi - \text{Tr}\left[\mathbf{K}_{\mathbf{U} \mathbf{U}}^{-1} \Phi \right] \right) \nonumber \\ 
& - \frac{1}{2\sigma^{2}} \text{Tr} \left[\mathbf{K}_{\mathbf{U} \mathbf{U}}^{-1} \Phi \mathbf{K}_{\mathbf{U} \mathbf{U}}^{-1}\left(\mathbf{M}_{:}\mathbf{M}_{:}^{\top} + \bm{\Sigma}^{\mathbf{U}_{:}} \right)\right],
\end{align}
where

\begin{align}
\Psi =\left\langle\mathbf{K}_{\mathbf{f}  \mathbf{U}}^{H}  \otimes \mathbf{K}_{\mathbf{f}  \mathbf{U}}^{X}\right\rangle_{q(\mathbf{H})} =  \left\langle\mathbf{K}_{\mathbf{f}  \mathbf{U}}^{H}\right\rangle_{q\left(\mathbf{H} \right)}  \otimes \mathbf{K}_{\mathbf{f} \mathbf{U}}^{X} = \Psi^{H}  \otimes \mathbf{K}_{\mathbf{f} \mathbf{U}}^{X},	
\end{align}

\begin{align}
\psi =\text{Tr}\left\langle\mathbf{K}_{\mathbf{f} \mathbf{f}}\right\rangle_{q\left(\mathbf{H}\right)} = \text{Tr}\left\langle\mathbf{K}_{\mathbf{f} \mathbf{f}}^{H} \otimes \mathbf{K}_{\mathbf{f} \mathbf{f}}^{X}\right\rangle_{q\left(\mathbf{H}\right)},	
\end{align}

\begin{align}
\Phi  &= \left\langle\mathbf{K}_{\mathbf{f}  \mathbf{U}}^{\top}\mathbf{K}_{\mathbf{f}  \mathbf{U}}\right\rangle_{q(\mathbf{H})}  = \left\langle\left(\mathbf{K}_{\mathbf{f}  \mathbf{U}}^{H}  \otimes \mathbf{K}_{\mathbf{f}  \mathbf{U}}^{X}\right)^{\top} \left(\mathbf{K}_{\mathbf{f} \mathbf{U}}^{H}  \otimes \mathbf{K}_{\mathbf{f} \mathbf{U}}^{X}\right)\right\rangle_{q(\mathbf{H})} \nonumber \\
& =  \Phi^{H} \otimes \left(\mathbf{K}_{\mathbf{f}  \mathbf{U}}^{X}\right)^{\top} \mathbf{K}_{\mathbf{f} \mathbf{U}}^{X}.
\end{align}

\subsection{Derivation of \texorpdfstring{$\mathcal{F}$}{\texttwoinferior} Given Different Input Datasets}
\label{appndx:HMOGPLV-Different_Outputs}
In this section, we show details for deriving $\mathcal{F}$ using different input datasets:
\begin{align} \mathcal{F} &= \mathbb{E}_{p \left(\mathbf{f} \mid \mathbf{U}_{:}, \mathbf{H}\right) q(\mathbf{U}_{:}) q(\mathbf{H})}\left[\log p\left(\mathbf{y} \mid \mathbf{f}, \mathbf{H}\right)\right] \nonumber \\ 
&= \int q(\mathbf{H}) \int q\left(\mathbf{U}_{:}\right) \underbrace{\int p \left(\mathbf{f} \mid \mathbf{U}_{:}, \mathbf{H}\right) \log p\left(\mathbf{y} \mid \mathbf{f}, \mathbf{H}\right)\text{d}\mathbf{f}}_{\mathcal{L}_{F}}\text{d}\mathbf{U}_{:}\text{d}\mathbf{H} \nonumber \\ 
&= \int q(\mathbf{H}) \underbrace{\int q\left(\mathbf{U}_{:}\right) \mathcal{L}_{F}\text{d}\mathbf{U}_{:}}_{\mathcal{L}_{U}}\text{d}\mathbf{H} \nonumber \\ &=\underbrace{\int q(\mathbf{H}) \mathcal{L}_{U}\text{d}\mathbf{H}}_{\mathcal{L}_{H}}. 
\end{align}

Now, we calculate $\mathcal{L}_{F}$: 
\begin{align} 
\mathcal{L}_{F} = & \int \prod_{d=1}^{D}p \left(\mathbf{f}_{d} \mid \mathbf{U}_{:}, \mathbf{H}\right) \log \prod_{d=1}^{D} p\left(\mathbf{y}_{d} \mid \mathbf{f}_{d}, \mathbf{H} \right)\text{d}\mathbf{f}_{d} \nonumber \\ 
= & \sum_{d=1}^{D}  \int p \left(\mathbf{f}_{d} \mid \mathbf{U}_{:}, \mathbf{H}\right) \log p\left(\mathbf{y}_{d} \mid \mathbf{f}_{d}, \mathbf{H} \right)\text{d}\mathbf{f}_{d} \nonumber \\
= & \sum_{d=1}^{D}  \left(\log \mathcal{N}\left(\mathbf{y}_{d} \mid \mathbf{K}_{\mathbf{f}_{d} \mathbf{U}} \mathbf{K}_{\mathbf{U} \mathbf{U}}^{-1} \mathbf{U}_{:}, \sigma_{d}^{2}\right) - \frac{1}{2\sigma_{d}^{2}} \text{Tr} \left[\mathbf{K}_{\mathbf{f}_{d} \mathbf{f}_{d}}-\mathbf{K}_{\mathbf{f}_{d} \mathbf{U}} \mathbf{K}_{\mathbf{U} \mathbf{U}}^{-1} \mathbf{K}_{\mathbf{f}_{d} \mathbf{U}}^{\top} \right]\right),
\end{align}
where $p\left(\mathbf{f}_{d} \mid \mathbf{U}_{:}, \mathbf{H}\right)=\mathcal{N}\left(\mathbf{f}_{d} \mid \mathbf{K}_{\mathbf{f}_{d} \mathbf{U}} \mathbf{K}_{\mathbf{U} \mathbf{U}}^{-1} \mathbf{U}_{:}, \mathbf{K}_{\mathbf{f}_{d} \mathbf{f}_{d}}-\mathbf{K}_{\mathbf{f}_{d} \mathbf{U}} \mathbf{K}_{\mathbf{U} \mathbf{U}}^{-1} \mathbf{K}_{\mathbf{f}_{d} \mathbf{U}}^{\top}\right)$. Then, we consider the $\mathcal{L}_{U}$:
\begin{align}
\mathcal{L}_{U} = & \int q\left(\mathbf{U}_{:}\right) \mathcal{L}_{F}\text{d}\mathbf{U}_{:} \nonumber \\
= & \int q\left(\mathbf{U}_{:}\right) \sum_{d=1}^{D}  \Bigg(\log \mathcal{N}\left(\mathbf{y}_{d} \mid \mathbf{K}_{\mathbf{f}_{d} \mathbf{U}} \mathbf{K}_{\mathbf{U} \mathbf{U}}^{-1} \mathbf{U}_{:}, \sigma_{d}^{2}\right)  \nonumber \\ 
& - \frac{1}{2\sigma_{d}^{2}}\text{Tr} \left[\mathbf{K}_{\mathbf{f}_{d} \mathbf{f}_{d}}-\mathbf{K}_{\mathbf{f}_{d} \mathbf{U}} \mathbf{K}_{\mathbf{U} \mathbf{U}}^{-1} \mathbf{K}_{\mathbf{f}_{d} \mathbf{U}}^{\top} \right]\Bigg)\text{d}\mathbf{U}_{:} \nonumber \\
= & \sum_{d=1}^{D}   \Big(\log \mathcal{N}\left(\mathbf{y}_{d} \mid \mathbf{K}_{\mathbf{f}_{d} \mathbf{U}} \mathbf{K}_{\mathbf{U} \mathbf{U}}^{-1} \mathbf{M}_{:}, \sigma_{d}^{2}\right) - \frac{1}{2\sigma_{d}^{2}} \text{Tr} \left[\mathbf{K}_{\mathbf{f}_{d} \mathbf{f}_{d}}-\mathbf{K}_{\mathbf{f}_{d} \mathbf{U}} \mathbf{K}_{\mathbf{U} \mathbf{U}}^{-1} \mathbf{K}_{\mathbf{f}_{d} \mathbf{U}}^{\top}\right] \nonumber \\
& - \frac{1}{2\sigma_{d}^{2}} \text{Tr} \left[\bm{\Sigma}^{\mathbf{U}_{:}} \mathbf{K}_{\mathbf{U} \mathbf{U}}^{-1} \mathbf{K}_{\mathbf{f}_{d} \mathbf{U}}^{\top}\mathbf{K}_{\mathbf{f}_{d} \mathbf{U}}\mathbf{K}_{\mathbf{U} \mathbf{U}}^{-1}\right]\Big),
\end{align}
where $q(\mathbf{U}_{:})=\mathcal{N}\left(\mathbf{U}_{:} \mid \mathbf{M}_{:}, \bm{\Sigma}^{\mathbf{U}_{:}}\right)$. 
Further, we obtain $\mathcal{L}_{H}$: 
\begin{align}
\mathcal{L}_{H}  = &\int q(\mathbf{H}) \mathcal{L}_{U}\text{d}\mathbf{H} \nonumber \\
= & \sum_{d=1}^{D}  \left\langle\log \mathcal{N}\left(\mathbf{y}_{d} \mid \mathbf{K}_{\mathbf{f}_{d} \mathbf{U}} \mathbf{K}_{\mathbf{U} \mathbf{U}}^{-1} \mathbf{M}_{:}, \sigma_{d}^{2}\right)\right\rangle_{q(\mathbf{h}_{d})} \nonumber \\
& - \frac{1}{2\sigma_{d}^{2}} \text{Tr} \left[\left\langle\mathbf{K}_{\mathbf{f}_{d} \mathbf{f}_{d}}\right\rangle_{q(\mathbf{h}_{d})} -\left\langle\mathbf{K}_{\mathbf{f}_{d} \mathbf{U}} \mathbf{K}_{\mathbf{U} \mathbf{U}}^{-1}\mathbf{K}_{\mathbf{f}_{d} \mathbf{U}}^{\top}\right\rangle_{q(\mathbf{h}_{d})}\right] \nonumber \\
&- \frac{1}{2\sigma_{d}^{2}} \text{Tr} \left[\bm{\Sigma}^{\mathbf{U}_{:}} \mathbf{K}_{\mathbf{U} \mathbf{U}}^{-1} \left\langle\mathbf{K}_{\mathbf{f}_{d} \mathbf{U}}^{\top}\mathbf{K}_{\mathbf{f}_{d} \mathbf{U}}\right\rangle_{q(\mathbf{h}_{d})}\mathbf{K}_{\mathbf{U} \mathbf{U}}^{-1}\right] \nonumber \\
= & \sum_{d=1}^{D}  \underbrace{-\frac{N_{d}R}{2} \log 2 \pi \sigma_{d}^{2}-\frac{1}{2 \sigma_{d}^{2}} \mathbf{y}_{d}^{\top} \mathbf{y}_{d}}_{\textbf{C}_{d}} + \frac{1}{\sigma_{d}^{2}} \mathbf{y}_{d}^{\top} \underbrace{ \left\langle\mathbf{K}_{\mathbf{f}_{d} \mathbf{U}}\right\rangle_{q(\mathbf{h}_{d})}}_{\Psi_{d}}\mathbf{K}_{\mathbf{U} \mathbf{U}}^{-1} \mathbf{M}_{:} \nonumber \\
& - \frac{1}{2 \sigma_{d}^{2}} \mathbf{M}_{:}^{\top}\mathbf{K}_{\mathbf{U} \mathbf{U}}^{-1} \underbrace{\left\langle\mathbf{K}_{\mathbf{f}_{d} \mathbf{U}}^{\top}\mathbf{K}_{\mathbf{f}_{d} \mathbf{U}}\right\rangle_{q(\mathbf{h}_{d})}}_{\Phi_{d}}\mathbf{K}_{\mathbf{U} \mathbf{U}}^{-1}\mathbf{M}_{:} - \frac{1}{2 \sigma_{d}^{2}} \underbrace{\text{Tr} \left\langle\mathbf{K}_{\mathbf{f}_{d} \mathbf{f}_{d}}\right\rangle_{q(\mathbf{h}_{d})}}_{\psi_{d}} \nonumber \\
& + \frac{1}{2\sigma_{d}^{2}} \text{Tr}\left[\mathbf{K}_{\mathbf{U} \mathbf{U}}^{-1} \underbrace{\left\langle\mathbf{K}_{\mathbf{f}_{d} \mathbf{U}}^{\top}\mathbf{K}_{\mathbf{f}_{d} \mathbf{U}}\right\rangle_{q(\mathbf{h}_{d})}}_{\Phi_{d}}\right]  \nonumber \\
& - \frac{1}{2\sigma_{d}^{2}} \text{Tr} \left[\bm{\Sigma}^{\mathbf{U}_{:}} \mathbf{K}_{\mathbf{U} \mathbf{U}}^{-1} \underbrace{\left\langle\mathbf{K}_{\mathbf{f}_{d} \mathbf{U}}^{\top}\mathbf{K}_{\mathbf{f}_{d} \mathbf{U}}\right\rangle_{q(\mathbf{h}_{d})}}_{\Phi_{d}}\mathbf{K}_{\mathbf{U} \mathbf{U}}^{-1}\right] \nonumber \\
= & \sum_{d=1}^{D}  \textbf{C}_{d} + \frac{1}{\sigma_{d}^{2}} \mathbf{y}_{d}^{\top} \Psi_{d}\mathbf{K}_{\mathbf{U} \mathbf{U}}^{-1} \mathbf{M}_{:} - \frac{1}{2 \sigma_{d}^{2}} \mathbf{M}_{:}^{\top}\mathbf{K}_{\mathbf{U} \mathbf{U}}^{-1} \Phi_{d}\mathbf{K}_{\mathbf{U} \mathbf{U}}^{-1}\mathbf{M}_{:} - \frac{1}{2 \sigma_{d}^{2}} \psi_{d} \nonumber \\
& + \frac{1}{2\sigma_{d}^{2}} \text{Tr}\left[\mathbf{K}_{\mathbf{U} \mathbf{U}}^{-1} \Phi_{d}\right] - \frac{1}{2\sigma_{d}^{2}} \text{Tr} \left[\bm{\Sigma}^{\mathbf{U}_{:}} \mathbf{K}_{\mathbf{U} \mathbf{U}}^{-1} \Phi_{d}\mathbf{K}_{\mathbf{U} \mathbf{U}}^{-1}\right] \nonumber \\
= & \sum_{d=1}^{D}  \textbf{C}_{d} + \frac{1}{\sigma_{d}^{2}} \mathbf{y}_{d}^{\top} \Psi_{d}\mathbf{K}_{\mathbf{U} \mathbf{U}}^{-1} \mathbf{M}_{:} -  \frac{1}{2 \sigma_{d}^{2}} \left( \psi_{d} - \text{Tr}\left[\mathbf{K}_{\mathbf{U} \mathbf{U}}^{-1} \Phi_{d}\right] \right) \nonumber \\
& - \frac{1}{2\sigma_{d}^{2}} \text{Tr} \left[\mathbf{K}_{\mathbf{U} \mathbf{U}}^{-1} \Phi_{d}\mathbf{K}_{\mathbf{U} \mathbf{U}}^{-1}\left(\mathbf{M}_{:}\mathbf{M}_{:}^{\top} + \bm{\Sigma}^{\mathbf{U}_{:}} \right)\right],
\end{align}
where $q(\mathbf{H}) = \prod_{d=1}^{D}q(\mathbf{h}_d)$ and
\begin{align}
\Psi_{d} =\left\langle\mathbf{K}_{\mathbf{f}_{d}  \mathbf{U}}^{H}  \otimes \mathbf{K}_{\mathbf{f}_{d}  \mathbf{U}}^{X}\right\rangle_{q(\mathbf{h}_{d})} =  \left\langle\mathbf{K}_{\mathbf{f}_{d}  \mathbf{U}}^{H}\right\rangle_{q\left(\mathbf{h}_{d} \right)}  \otimes \mathbf{K}_{\mathbf{f}_{d} \mathbf{U}}^{X} = \Psi^{H}_{d}  \otimes \mathbf{K}_{\mathbf{f}_{d} \mathbf{U}}^{X}
\end{align}
\begin{align}
\psi_{d} =\text{Tr}\left\langle\mathbf{K}_{\mathbf{f}_{d} \mathbf{f}_{d}}\right\rangle_{q\left(\mathbf{h}_{d}\right)} = \text{Tr}\left\langle\mathbf{K}_{\mathbf{f}_{d} \mathbf{f}_{d}}^{H} \otimes \mathbf{K}_{\mathbf{f}_{d} \mathbf{f}_{d}}^{X}\right\rangle_{q\left(\mathbf{h}_{d}\right)},
\end{align}
\begin{align}
\Phi_{d}  = \left\langle\mathbf{K}_{\mathbf{f}_{d}  \mathbf{U}}^{\top}\mathbf{K}_{\mathbf{f}_{d}   \mathbf{U}}\right\rangle_{q(\mathbf{h}_{d})} & = \left\langle\left(\mathbf{K}_{\mathbf{f}_{d}  \mathbf{U}}^{H}  \otimes \mathbf{K}_{\mathbf{f}_{d}  \mathbf{U}}^{X}\right)^{\top} \left(\mathbf{K}_{\mathbf{f}_{d} \mathbf{U}}^{H}  \otimes \mathbf{K}_{\mathbf{f}_{d} \mathbf{U}}^{X}\right)\right\rangle_{q(\mathbf{h}_{d})} \nonumber \\ 
& = \left\langle\left(\mathbf{K}_{\mathbf{f}_{d}  \mathbf{U}}^{H}\right)^{\top}\mathbf{K}_{\mathbf{f}_{d} \mathbf{U}}^{H}\right\rangle_{q(\mathbf{h}_{d})}  \otimes \left(\mathbf{K}_{\mathbf{f}_{d}  \mathbf{U}}^{X}\right)^{\top}         \mathbf{K}_{\mathbf{f}_{d} \mathbf{U}}^{X} \nonumber \\
& =  \Phi_{d}^{H} \otimes \left(\mathbf{K}_{\mathbf{f}_{d}  \mathbf{U}}^{X}\right)^{\top} \mathbf{K}_{\mathbf{f}_{d} \mathbf{U}}^{X}.
\end{align}

\section{More Efficient Formulations}
\label{appndx:HMOGPLV-effcient}

In this subsection, we reduce the computational complexity by exploiting the Kronecker product decomposition. To fully utilise its properties, we assume there is a Kronecker product decomposition of the covariance matrix of $q(\mathbf{U}_{:})$, $\bm{\Sigma}^{\mathbf{U}_{:}} = \bm{\Sigma}^{H_{:}} \otimes \bm{\Sigma}^{X_{:}}$ and this format can reduce variational parameters from $M_{\mathbf{X}}^{2} M_{\mathbf{H}}^{2}$ to $M_{\mathbf{X}}^{2}+M_{\mathbf{H}}^{2}$ in $q(\mathbf{U}_{:})$. We also reformulate $\Phi$, $\Psi$, $\psi$ as

\begin{align}
\Phi  &= \left\langle\mathbf{K}_{\mathbf{f}  \mathbf{U}}^{\top}\mathbf{K}_{\mathbf{f}  \mathbf{U}}\right\rangle_{q(\mathbf{H})} = \Phi^{H} \otimes \Phi^{X},\\
\Phi^{H} &= \left\langle\left(\mathbf{K}_{\mathbf{f}  \mathbf{U}}^{H}\right)^{\top}\mathbf{K}_{\mathbf{f} \mathbf{U}}^{H}\right\rangle_{q(\mathbf{H})},\\
\Phi^{X} &= \left(\mathbf{K}_{\mathbf{f}  \mathbf{U}}^{X}\right)^{\top} \mathbf{K}_{\mathbf{f} \mathbf{U}}^{X},\\
\Psi &=\left\langle\mathbf{K}_{\mathbf{f}  \mathbf{U}}^{H}  \otimes \mathbf{K}_{\mathbf{f}  \mathbf{U}}^{X}\right\rangle_{q(\mathbf{H})} =  \left\langle\mathbf{K}_{\mathbf{f}  \mathbf{U}}^{H}\right\rangle_{q\left(\mathbf{H} \right)}  \otimes \mathbf{K}_{\mathbf{f} \mathbf{U}}^{X} = \Psi^{H}  \otimes \mathbf{K}_{\mathbf{f} \mathbf{U}}^{X},\\
\psi & =\text{Tr}\left\langle\mathbf{K}_{\mathbf{f} \mathbf{f}}\right\rangle_{q\left(\mathbf{H}\right)} = \text{Tr}\left\langle\mathbf{K}_{\mathbf{f} \mathbf{f}}^{H} \otimes \mathbf{K}_{\mathbf{f} \mathbf{f}}^{X}\right\rangle_{q\left(\mathbf{H}\right)}.
\end{align}
Using the property of the Kronecker product decomposition, we obtain a new format of the lower bound (for more detail see Section \ref{appndx:HMOGPLV-efficient-sameinput}):
\begin{align} 
\mathcal{F}=  &-\frac{NDR}{2} \log 2 \pi \sigma^{2}-\frac{1}{2 \sigma^{2}} \mathbf{y}^{\top} \mathbf{y} \nonumber \\ 
& - \frac{1}{2 \sigma^{2}} \text{Tr}\left( \mathbf{M}^{\top} \left(\mathbf{K}_{\mathbf{U} \mathbf{U}}^{X}\right)^{-1}\Phi^{X} \left(\mathbf{K}_{\mathbf{U} \mathbf{U}}^{X}\right)^{-1} \mathbf{M} \left(\mathbf{K}_{\mathbf{U} \mathbf{U}}^{H}\right)^{-1}\Phi^{H}\left(\mathbf{K}_{\mathbf{U} \mathbf{U}}^{H}\right)^{-1} \right) \nonumber\\
& - \frac{1}{2 \sigma^{2}} \text{Tr} \left(\left(\mathbf{K}_{\mathbf{U} \mathbf{U}}^{H}\right)^{-1}\Phi^{H}\left(\mathbf{K}_{\mathbf{U} \mathbf{U}}^{H}\right)^{-1} \bm{\Sigma}^{H_{:}}\right)   \text{Tr} \left(\left(\mathbf{K}_{\mathbf{U} \mathbf{U}}^{X}\right)^{-1}\Phi^{X} \left(\mathbf{K}_{\mathbf{U} \mathbf{U}}^{X}\right)^{-1}\bm{\Sigma}^{X_{:}}\right) \nonumber\\
& +\frac{1}{\sigma^{2}} \mathbf{y}^{\top} \left(\mathbf{K}_{\mathbf{f} \mathbf{U}}^{X} \left(\mathbf{K}_{\mathbf{U} \mathbf{U}}^{X}\right)^{-1} \mathbf{M} \left(\mathbf{K}_{\mathbf{U} \mathbf{U}}^{H}\right)^{-1} \left(\Psi^{H}\right)^{\top} \right)_{:} -\frac{1}{2 \sigma^{2}}\psi \nonumber \\
& + \frac{1}{2\sigma^{2}} \text{Tr}\left(\left(\mathbf{K}_{\mathbf{U} \mathbf{U}}^{H}\right)^{-1} \Phi^{H}\right) \text{Tr}\left( \left(\mathbf{K}_{\mathbf{U} \mathbf{U}}^{X}\right)^{-1} \Phi^{X}  \right).
\end{align}
Similarly, the KL-divergence between $q(\mathbf{U}_{:})$ and $p(\mathbf{U}_{:})$ can also benefit from the above decomposition (see Section \ref{appndx:HMOGPLV-efficient-sameinput} for more detail):
\begin{align}
\operatorname{KL}\left\{q\left(\mathbf{U}_{:}\right) \mid p\left(\mathbf{U}_{:}\right)\right\} = & \frac{1}{2}\Bigg(M_{\mathbf{X}} \log \frac{\left|\mathbf{K}_{\mathbf{U} \mathbf{U}}^{H}\right|}{\left|\bm{\Sigma}^{H_{:}}\right|}+M_{\mathbf{H}} \log \frac{\left|\mathbf{K}_{\mathbf{U} \mathbf{U}}^{X}\right|}{\left|\bm{\Sigma}^{X_{:}}\right|} \nonumber \\
& +\text{Tr} \left(\mathbf{M}^{\top}\left(\mathbf{K}_{\mathbf{U} \mathbf{U}}^{X}\right)^{-1} \mathbf{M}\left(\mathbf{K}_{\mathbf{U} \mathbf{U}}^{H}\right)^{-1}\right) \nonumber \\
&+\text{Tr} \left(\left(\mathbf{K}_{\mathbf{U} \mathbf{U}}^{H}\right)^{-1} \bm{\Sigma}^{H_{:}}\right) \text{Tr} \left(\left(\mathbf{K}_{\mathbf{U} \mathbf{U}}^{X}\right)^{-1} \bm{\Sigma}^{X_{:}}\right)-M_{\mathbf{H}} M_{\mathbf{X}}\Bigg).
\end{align}
The computational complexity of $\mathcal{L}$ is led by the product $\left(\mathbf{K}_{\mathbf{f}  \mathbf{U}}^{H}\right)^{\top}\mathbf{K}_{\mathbf{f} \mathbf{U}}^{H}$ and $\left(\mathbf{K}_{\mathbf{f}  \mathbf{U}}^{X}\right)^{\top} \mathbf{K}_{\mathbf{f} \mathbf{U}}^{X}$ with a cost of $\mathcal{O}\left(DM_{\mathbf{H}}^{2}\right)$ and $\mathcal{O}\left(NRM_{\mathbf{X}}^{2}\right)$, respectively, which is more efficient than the original formulation. Further, we can extend the lower bound with by using mini-batches to improve its scalability.

Besides, we can also reduce the computational complexity in $\mathcal{F}$ and Kullback–Leibler divergence by taking the advantage of the Kronecker product decomposition.

\subsection{Datasets with Common Inputs}
\label{appndx:HMOGPLV-efficient-sameinput}

In this section, given the same input datasets, we re-define $\mathcal{F}$ and Kullback–Leibler divergence by using the Kronecker product decomposition, such that:
\begin{align} 
\mathcal{F}=  &-\frac{NDR}{2} \log 2 \pi \sigma^{2}-\frac{1}{2 \sigma^{2}} \mathbf{y}^{\top} \mathbf{y} \nonumber\\ 
& - \frac{1}{2 \sigma^{2}} \text{Tr}\left(\left(\mathbf{K}_{\mathbf{U} \mathbf{U}}^{H}  \otimes \mathbf{K}_{\mathbf{U} \mathbf{U}}^{X}\right)^{-1} \left(\Phi^{H} \otimes \Phi^{X}\right) \left(\mathbf{K}_{\mathbf{U} \mathbf{U}}^{H}  \otimes \mathbf{K}_{\mathbf{U} \mathbf{U}}^{X}\right)^{-1}\mathbf{M}_{:} \mathbf{M}_{:}^{\top}\right) \nonumber \\
& - \frac{1}{2 \sigma^{2}} \text{Tr}\left(\left(\mathbf{K}_{\mathbf{U} \mathbf{U}}^{H}  \otimes \mathbf{K}_{\mathbf{U} \mathbf{U}}^{X}\right)^{-1} \left(\Phi^{H} \otimes \Phi^{X}\right) \left(\mathbf{K}_{\mathbf{U} \mathbf{U}}^{H}  \otimes \mathbf{K}_{\mathbf{U} \mathbf{U}}^{X}\right)^{-1}\left(\bm{\Sigma}^{H_{:}} \otimes \bm{\Sigma}^{X_{:}}\right)\right) \nonumber \\
& +\frac{1}{\sigma^{2}} \mathbf{y}^{\top} \left(\Psi^{H}  \otimes \mathbf{K}_{\mathbf{f} \mathbf{U}}^{X}\right) \left(\mathbf{K}_{\mathbf{U} \mathbf{U}}^{H}  \otimes \mathbf{K}_{\mathbf{U} \mathbf{U}}^{X}\right)^{-1} \mathbf{M}_{:} -\frac{1}{2 \sigma^{2}}\psi \nonumber \\
& + \frac{1}{2 \sigma^{2}}\left(\text{Tr}\left(\left(\mathbf{K}_{\mathbf{U} \mathbf{U}}^{H}  \otimes \mathbf{K}_{\mathbf{U} \mathbf{U}}^{X}\right)^{-1} \left(\Phi^{H} \otimes \Phi^{X}\right) \right)\right)\nonumber \\
=  &-\frac{NDR}{2} \log 2 \pi \sigma^{2}-\frac{1}{2 \sigma^{2}} \mathbf{y}^{\top} \mathbf{y} \nonumber \\ 
& - \frac{1}{2 \sigma^{2}} \text{Tr}\left( \left(\left(\mathbf{K}_{\mathbf{U} \mathbf{U}}^{H}\right)^{-1}\Phi^{H}\left(\mathbf{K}_{\mathbf{U} \mathbf{U}}^{H}\right)^{-1}\right)  \otimes \left(\left(\mathbf{K}_{\mathbf{U} \mathbf{U}}^{X}\right)^{-1}\Phi^{X} \left(\mathbf{K}_{\mathbf{U} \mathbf{U}}^{X}\right)^{-1}\right) \mathbf{M}_{:} \mathbf{M}_{:}^{\top}\right) \nonumber \\
& - \frac{1}{2 \sigma^{2}} \text{Tr}\left(\left(\left(\mathbf{K}_{\mathbf{U} \mathbf{U}}^{H}\right)^{-1}\Phi^{H}\left(\mathbf{K}_{\mathbf{U} \mathbf{U}}^{H}\right)^{-1} \bm{\Sigma}^{H_{:}}\right)  \otimes \left(\left(\mathbf{K}_{\mathbf{U} \mathbf{U}}^{X}\right)^{-1}\Phi^{X} \left(\mathbf{K}_{\mathbf{U} \mathbf{U}}^{X}\right)^{-1}\bm{\Sigma}^{X_{:}}\right)\right) \nonumber \\
& +\frac{1}{\sigma^{2}} \mathbf{y}^{\top} \left(\Psi^{H} \left(\mathbf{K}_{\mathbf{U} \mathbf{U}}^{H}\right)^{-1}   \otimes \mathbf{K}_{\mathbf{f} \mathbf{U}}^{X} \left(\mathbf{K}_{\mathbf{U} \mathbf{U}}^{X}\right)^{-1}\right) \mathbf{M}_{:} -\frac{1}{2 \sigma^{2}}\psi \nonumber \\
& + \frac{1}{2 \sigma^{2}}\left(\text{Tr}\left(\left(\mathbf{K}_{\mathbf{U} \mathbf{U}}^{H}\right)^{-1} \Phi^{H} \otimes \left(\mathbf{K}_{\mathbf{U} \mathbf{U}}^{X}\right)^{-1} \Phi^{X}  \right)\right) \nonumber \\
=  &-\frac{NDR}{2} \log 2 \pi \sigma^{2}-\frac{1}{2 \sigma^{2}} \mathbf{y}^{\top} \mathbf{y} \nonumber \\ 
& - \frac{1}{2 \sigma^{2}} \text{Tr}\left( \mathbf{M}^{\top} \left(\mathbf{K}_{\mathbf{U} \mathbf{U}}^{X}\right)^{-1}\Phi^{X} \left(\mathbf{K}_{\mathbf{U} \mathbf{U}}^{X}\right)^{-1} \mathbf{M} \left(\mathbf{K}_{\mathbf{U} \mathbf{U}}^{H}\right)^{-1}\Phi^{H}\left(\mathbf{K}_{\mathbf{U} \mathbf{U}}^{H}\right)^{-1} \right) \nonumber \\
& - \frac{1}{2 \sigma^{2}} \text{Tr} \left(\left(\mathbf{K}_{\mathbf{U} \mathbf{U}}^{H}\right)^{-1}\Phi^{H}\left(\mathbf{K}_{\mathbf{U} \mathbf{U}}^{H}\right)^{-1} \bm{\Sigma}^{H_{:}}\right)   \text{Tr} \left(\left(\mathbf{K}_{\mathbf{U} \mathbf{U}}^{X}\right)^{-1}\Phi^{X} \left(\mathbf{K}_{\mathbf{U} \mathbf{U}}^{X}\right)^{-1}\bm{\Sigma}^{X_{:}}\right) \nonumber \\
& +\frac{1}{\sigma^{2}} \mathbf{y}^{\top} \left(\mathbf{K}_{\mathbf{f} \mathbf{U}}^{X} \left(\mathbf{K}_{\mathbf{U} \mathbf{U}}^{X}\right)^{-1} \mathbf{M} \left(\mathbf{K}_{\mathbf{U} \mathbf{U}}^{H}\right)^{-1} \left(\Psi^{H}\right)^{\top} \right)_{:} -\frac{1}{2 \sigma^{2}}\psi \nonumber\\
& + \frac{1}{2\sigma^{2}} \text{Tr}\left(\left(\mathbf{K}_{\mathbf{U} \mathbf{U}}^{H}\right)^{-1} \Phi^{H}\right) \text{Tr}\left( \left(\mathbf{K}_{\mathbf{U} \mathbf{U}}^{X}\right)^{-1} \Phi^{X}  \right).
\end{align}
We also assume there is a Kronecker product decomposition of the covariance matrix of $q(\mathbf{U}_{:})$, $\bm{\Sigma}^{\mathbf{U}_{:}} = \bm{\Sigma}^{H_{:}} \otimes \bm{\Sigma}^{X_{:}}$ so the KL-divergence between $q(\mathbf{U}_{:})$ and $p(\mathbf{U}_{:})$ can also take advantage of the decomposition:
\begin{align}
& \text{KL}\left\{q\left(\mathbf{U}_{:}\right) \mid p\left(\mathbf{U}_{:}\right)\right\} \nonumber \\
= & \frac{1}{2}\Bigg(\log \left|\mathbf{K}_{\mathbf{U} \mathbf{U}}^{H}  \otimes \mathbf{K}_{\mathbf{U} \mathbf{U}}^{X}\left(\boldsymbol{\Sigma}^{H_{:}} \otimes \bm{\Sigma}^{X_{:}}\right)^{-1}\right| \nonumber \\
& + \text{Tr}\left(\left(\mathbf{K}_{\mathbf{U} \mathbf{U}}^{H}  \otimes \mathbf{K}_{\mathbf{U} \mathbf{U}}^{X}\right)^{-1}\left(\mathbf{M}_{:} \mathbf{M}_{:}^{\top}+\bm{\Sigma}^{H_{:}} \otimes \bm{\Sigma}^{X_{:}}-\left(\mathbf{K}_{\mathbf{U} \mathbf{U}}^{H}  \otimes \mathbf{K}_{\mathbf{U} \mathbf{U}}^{X}\right)\right)\right)\Bigg) \nonumber \\
= &  \frac{1}{2}\Bigg(\log \left|\mathbf{K}_{\mathbf{U} \mathbf{U}}^{H} \left(\boldsymbol{\Sigma}^{H_{:}}\right)^{-1}  \otimes \mathbf{K}_{\mathbf{U} \mathbf{U}}^{X}\left(\bm{\Sigma}^{X_{:}}\right)^{-1}\right| \nonumber \\
& + \text{Tr}\left(\left(\mathbf{K}_{\mathbf{U} \mathbf{U}}^{H}  \otimes \mathbf{K}_{\mathbf{U} \mathbf{U}}^{X}\right)^{-1}\left(\mathbf{M}_{:} \mathbf{M}_{:}^{\top}+\bm{\Sigma}^{H_{:}} \otimes \bm{\Sigma}^{X_{:}} \right)\right) - M_{H} M_{X} \Bigg) \nonumber \\
= &  \frac{1}{2}\Bigg(M_{X}\log \left|\mathbf{K}_{\mathbf{U} \mathbf{U}}^{H} \left(\boldsymbol{\Sigma}^{H_{:}}\right)^{-1}\right| + M_{H}\log \left|\mathbf{K}_{\mathbf{U} \mathbf{U}}^{X}\left(\bm{\Sigma}^{X_{:}}\right)^{-1}\right| \nonumber \\
& + \text{Tr} \left(\mathbf{M}^{\top}\left(\mathbf{K}_{\mathbf{U} \mathbf{U}}^{X}\right)^{-1} \mathbf{M}\left(\mathbf{K}_{\mathbf{U} \mathbf{U}}^{H}\right)^{-1}\right) \nonumber \\
& + \text{Tr}\left(\left(\mathbf{K}_{\mathbf{U} \mathbf{U}}^{H}\right)^{-1} \bm{\Sigma}^{H_{:}} \right)  \text{Tr}\left( \left(\mathbf{K}_{\mathbf{U} \mathbf{U}}^{X}\right)^{-1} \bm{\Sigma}^{X_{:}}\right) - M_{H} M_{X} \Bigg) \nonumber \\
= & \frac{1}{2}\left(M_{X} \log \frac{\left|\mathbf{K}_{\mathbf{U} \mathbf{U}}^{H}\right|}{\left|\bm{\Sigma}^{H_{:}}\right|}+M_{H} \log \frac{\left|\mathbf{K}_{\mathbf{U} \mathbf{U}}^{X}\right|}{\left|\bm{\Sigma}^{X_{:}}\right|}+\text{Tr} \left(\mathbf{M}^{\top}\left(\mathbf{K}_{\mathbf{U} \mathbf{U}}^{X}\right)^{-1} \mathbf{M}\left(\mathbf{K}_{\mathbf{U} \mathbf{U}}^{H}\right)^{-1}\right)\right. \nonumber\\
&\left.+\text{Tr} \left(\left(\mathbf{K}_{\mathbf{U} \mathbf{U}}^{H}\right)^{-1} \bm{\Sigma}^{H_{:}}\right) \text{Tr} \left(\left(\mathbf{K}_{\mathbf{U} \mathbf{U}}^{X}\right)^{-1} \bm{\Sigma}^{X_{:}}\right)-M_{H} M_{X}\right).
\end{align}

\subsection{Datasets with Different Inputs}
\label{appndx:HMOGPLV-efficient-differentinput}

As for common input datasets, we reformulate $\Phi_{d}$, $\Psi_{d}$, $\psi_{d}$ as
\begin{align}
\Phi_{d} & = \left\langle\mathbf{K}_{\mathbf{f}_{d}  \mathbf{U}}^{\top}\mathbf{K}_{\mathbf{f}_{d}   \mathbf{U}}\right\rangle_{q(\mathbf{h}_{d})}  = \left\langle\left(\mathbf{K}_{\mathbf{f}_{d}  \mathbf{U}}^{H}  \otimes \mathbf{K}_{\mathbf{f}_{d}  \mathbf{U}}^{X}\right)^{\top} \left(\mathbf{K}_{\mathbf{f}_{d} \mathbf{U}}^{H}  \otimes \mathbf{K}_{\mathbf{f}_{d} \mathbf{U}}^{X}\right)\right\rangle_{q(\mathbf{h}_{d})} \nonumber \\
& =  \Phi_{d}^{H} \otimes \Phi_{d}^{X},\\
\Phi_{d}^{H} &  =  \left\langle\left(\mathbf{K}_{\mathbf{f}_{d}  \mathbf{U}}^{H}\right)^{\top}\mathbf{K}_{\mathbf{f}_{d} \mathbf{U}}^{H}\right\rangle_{q(\mathbf{h}_{d})},\\ 
\Phi_{d}^{X} & =  \left(\mathbf{K}_{\mathbf{f}_{d}  \mathbf{U}}^{X}\right)^{\top} \mathbf{K}_{\mathbf{f}_{d} \mathbf{U}}^{X},\\
\Psi_{d} &=\left\langle\mathbf{K}_{\mathbf{f}_{d}  \mathbf{U}}^{H}  \otimes \mathbf{K}_{\mathbf{f}_{d}  \mathbf{U}}^{X}\right\rangle_{q(\mathbf{h}_{d})} =  \left\langle\mathbf{K}_{\mathbf{f}_{d}  \mathbf{U}}^{H}\right\rangle_{q\left(\mathbf{h}_{d} \right)}  \otimes \mathbf{K}_{\mathbf{f}_{d} \mathbf{U}}^{X} = \Psi^{H}_{d}  \otimes \mathbf{K}_{\mathbf{f}_{d} \mathbf{U}}^{X},\\
\psi_{d} & =\text{Tr}\left\langle\mathbf{K}_{\mathbf{f}_{d} \mathbf{f}_{d}}\right\rangle_{q\left(\mathbf{h}_{d}\right)} = \text{Tr}\left\langle\mathbf{K}_{\mathbf{f}_{d} \mathbf{f}_{d}}^{H} \otimes \mathbf{K}_{\mathbf{f}_{d} \mathbf{f}_{d}}^{X}\right\rangle_{q\left(\mathbf{h}_{d}\right)}.
\end{align}
We also reduce the computational complexity by using the property of the Kronecker product decomposition (for more detail, see Section \ref{appndx:HMOGPLV-efficient-differentinput}):
\begin{align} 
\mathcal{F} =  & \sum_{d=1}^{D} -\frac{N_{d}R}{2} \log 2 \pi \sigma_{d}^{2}-\frac{1}{2 \sigma_{d}^{2}} \mathbf{y}_{d}^{\top} \mathbf{y}_{d} \nonumber \\ 
& - \frac{1}{2 \sigma_{d}^{2}} \text{Tr}\left( \mathbf{M}^{\top} \left(\mathbf{K}_{\mathbf{U} \mathbf{U}}^{X}\right)^{-1}\Phi^{X}_{d} \left(\mathbf{K}_{\mathbf{U} \mathbf{U}}^{X}\right)^{-1} \mathbf{M} \left(\mathbf{K}_{\mathbf{U} \mathbf{U}}^{H}\right)^{-1}\Phi^{H}_{d}\left(\mathbf{K}_{\mathbf{U} \mathbf{U}}^{H}\right)^{-1} \right) \nonumber \\
& - \frac{1}{2 \sigma_{d}^{2}} \text{Tr} \left(\left(\mathbf{K}_{\mathbf{U} \mathbf{U}}^{H}\right)^{-1}\Phi^{H}_{d}\left(\mathbf{K}_{\mathbf{U} \mathbf{U}}^{H}\right)^{-1} \bm{\Sigma}^{H_{:}}\right)   \text{Tr} \left(\left(\mathbf{K}_{\mathbf{U} \mathbf{U}}^{X}\right)^{-1}\Phi^{X}_{d} \left(\mathbf{K}_{\mathbf{U} \mathbf{U}}^{X}\right)^{-1}\bm{\Sigma}^{X_{:}}\right) \nonumber \\
& +\frac{1}{\sigma_{d}^{2}} \mathbf{y}^{\top}_{d} \left(\mathbf{K}_{\mathbf{f}_{d} \mathbf{U}}^{X} \left(\mathbf{K}_{\mathbf{U} \mathbf{U}}^{X}\right)^{-1} \mathbf{M} \left(\mathbf{K}_{\mathbf{U} \mathbf{U}}^{H}\right)^{-1} \left(\Psi^{H}_{d}\right)^{\top} \right)_{:} -\frac{1}{2 \sigma_{d}^{2}}\psi_{d} \nonumber \\
& + \frac{1}{2\sigma_{d}^{2}} \text{Tr}\left(\left(\mathbf{K}_{\mathbf{U} \mathbf{U}}^{H}\right)^{-1} \Phi^{H}_{d}\right) \text{Tr}\left( \left(\mathbf{K}_{\mathbf{U} \mathbf{U}}^{X}\right)^{-1} \Phi^{X}_{d}  \right).
\end{align}
The computational complexity of the lower bound is mainly controlled by $\left(\mathbf{K}_{\mathbf{f}_{d}  \mathbf{U}}^{X}\right)^{\top}$ $\times$ $\mathbf{K}_{\mathbf{f}_{d} \mathbf{U}}^{X}$ with $\mathcal{O}\left(N_{d}RM_{\mathbf{X}}^{2}\right)$. We also can extend $\mathcal{L} $ by applying the mini-bath method to improve scalability of our model.

In this section, given the different input datasets, we re-define $\mathcal{F}$ using the Kronecker product decomposition.
\begin{align} 
\mathcal{F}=  & \sum_{d=1}^{D} -\frac{N_{d}R}{2} \log 2 \pi \sigma_{d}^{2}-\frac{1}{2 \sigma_{d}^{2}} \mathbf{y}_{d}^{\top} \mathbf{y}_{d} \nonumber \\ 
& - \frac{1}{2 \sigma_{d}^{2}} \text{Tr}\left(\left(\mathbf{K}_{\mathbf{U} \mathbf{U}}^{H}  \otimes \mathbf{K}_{\mathbf{U} \mathbf{U}}^{X}\right)^{-1} \left(\Phi^{H}_{d} \otimes \Phi^{X}_{d}\right) \left(\mathbf{K}_{\mathbf{U} \mathbf{U}}^{H}  \otimes \mathbf{K}_{\mathbf{U} \mathbf{U}}^{X}\right)^{-1}\mathbf{M}_{:} \mathbf{M}_{:}^{\top}\right) \nonumber\\
& - \frac{1}{2 \sigma_{d}^{2}} \text{Tr}\left(\left(\mathbf{K}_{\mathbf{U} \mathbf{U}}^{H}  \otimes \mathbf{K}_{\mathbf{U} \mathbf{U}}^{X}\right)^{-1} \left(\Phi^{H}_{d} \otimes \Phi^{X}_{d}\right) \left(\mathbf{K}_{\mathbf{U} \mathbf{U}}^{H}  \otimes \mathbf{K}_{\mathbf{U} \mathbf{U}}^{X}\right)^{-1}\left(\bm{\Sigma}^{H_{:}} \otimes \bm{\Sigma}^{X_{:}}\right)\right) \nonumber \\
& +\frac{1}{\sigma_{d}^{2}} \mathbf{y}^{\top}_{d} \left(\Psi^{H}_{d}  \otimes \mathbf{K}_{\mathbf{f}_{d} \mathbf{U}}^{X}\right) \left(\mathbf{K}_{\mathbf{U} \mathbf{U}}^{H}  \otimes \mathbf{K}_{\mathbf{U} \mathbf{U}}^{X}\right)^{-1} \mathbf{M}_{:} -\frac{1}{2 \sigma_{d}^{2}}\psi_{d}\nonumber \\
& + \frac{1}{2 \sigma_{d}^{2}}\left(\text{Tr}\left(\left(\mathbf{K}_{\mathbf{U} \mathbf{U}}^{H}  \otimes \mathbf{K}_{\mathbf{U} \mathbf{U}}^{X}\right)^{-1} \left(\Phi^{H}_{d} \otimes \Phi^{X}_{d}\right) \right)\right) \nonumber
\end{align}
\begin{align}
=  & \sum_{d=1}^{D}  -\frac{N_{d}R}{2} \log 2 \pi \sigma_{d}^{2}-\frac{1}{2 \sigma_{d}^{2}} \mathbf{y}_{d}^{\top} \mathbf{y}_{d} \nonumber\\ 
& - \frac{1}{2 \sigma_{d}^{2}} \text{Tr}\left( \left(\left(\mathbf{K}_{\mathbf{U} \mathbf{U}}^{H}\right)^{-1}\Phi^{H}_{d}\left(\mathbf{K}_{\mathbf{U} \mathbf{U}}^{H}\right)^{-1}\right)  \otimes \left(\left(\mathbf{K}_{\mathbf{U} \mathbf{U}}^{X}\right)^{-1}\Phi^{X}_{d} \left(\mathbf{K}_{\mathbf{U} \mathbf{U}}^{X}\right)^{-1}\right) \mathbf{M}_{:} \mathbf{M}_{:}^{\top}\right) \nonumber \\
& - \frac{1}{2 \sigma_{d}^{2}} \text{Tr}\left(\left(\left(\mathbf{K}_{\mathbf{U} \mathbf{U}}^{H}\right)^{-1}\Phi^{H}_{d}\left(\mathbf{K}_{\mathbf{U} \mathbf{U}}^{H}\right)^{-1} \bm{\Sigma}^{H_{:}}\right)  \otimes \left(\left(\mathbf{K}_{\mathbf{U} \mathbf{U}}^{X}\right)^{-1}\Phi^{X}_{d} \left(\mathbf{K}_{\mathbf{U} \mathbf{U}}^{X}\right)^{-1}\bm{\Sigma}^{X_{:}}\right)\right) \nonumber \\
& +\frac{1}{\sigma_{d}^{2}} \mathbf{y}^{\top}_{d} \left(\Psi^{H}_{d} \left(\mathbf{K}_{\mathbf{U} \mathbf{U}}^{H}\right)^{-1}   \otimes \mathbf{K}_{\mathbf{f}_{d} \mathbf{U}}^{X} \left(\mathbf{K}_{\mathbf{U} \mathbf{U}}^{X}\right)^{-1}\right) \mathbf{M}_{:} -\frac{1}{2 \sigma_{d}^{2}}\psi_{d}\nonumber \\
& + \frac{1}{2 \sigma_{d}^{2}}\left(\text{Tr}\left(\left(\mathbf{K}_{\mathbf{U} \mathbf{U}}^{H}\right)^{-1} \Phi^{H}_{d} \otimes \left(\mathbf{K}_{\mathbf{U} \mathbf{U}}^{X}\right)^{-1} \Phi^{X}_{d}  \right)\right)\nonumber \\
=  & \sum_{d=1}^{D} -\frac{N_{d}R}{2} \log 2 \pi \sigma_{d}^{2}-\frac{1}{2 \sigma_{d}^{2}} \mathbf{y}_{d}^{\top} \mathbf{y}_{d}\nonumber \\ 
& - \frac{1}{2 \sigma_{d}^{2}} \text{Tr}\left( \mathbf{M}^{\top} \left(\mathbf{K}_{\mathbf{U} \mathbf{U}}^{X}\right)^{-1}\Phi^{X}_{d} \left(\mathbf{K}_{\mathbf{U} \mathbf{U}}^{X}\right)^{-1} \mathbf{M} \left(\mathbf{K}_{\mathbf{U} \mathbf{U}}^{H}\right)^{-1}\Phi^{H}_{d}\left(\mathbf{K}_{\mathbf{U} \mathbf{U}}^{H}\right)^{-1} \right) \nonumber \\
& - \frac{1}{2 \sigma_{d}^{2}} \text{Tr} \left(\left(\mathbf{K}_{\mathbf{U} \mathbf{U}}^{H}\right)^{-1}\Phi^{H}_{d}\left(\mathbf{K}_{\mathbf{U} \mathbf{U}}^{H}\right)^{-1} \bm{\Sigma}^{H_{:}}\right)   \text{Tr} \left(\left(\mathbf{K}_{\mathbf{U} \mathbf{U}}^{X}\right)^{-1}\Phi^{X}_{d} \left(\mathbf{K}_{\mathbf{U} \mathbf{U}}^{X}\right)^{-1}\bm{\Sigma}^{X_{:}}\right) \nonumber \\
& +\frac{1}{\sigma_{d}^{2}} \mathbf{y}^{\top}_{d} \left(\mathbf{K}_{\mathbf{f}_{d} \mathbf{U}}^{X} \left(\mathbf{K}_{\mathbf{U} \mathbf{U}}^{X}\right)^{-1} \mathbf{M} \left(\mathbf{K}_{\mathbf{U} \mathbf{U}}^{H}\right)^{-1} \left(\Psi^{H}_{d}\right)^{\top} \right)_{:} -\frac{1}{2 \sigma_{d}^{2}}\psi_{d}\nonumber \\
& + \frac{1}{2\sigma_{d}^{2}} \text{Tr}\left(\left(\mathbf{K}_{\mathbf{U} \mathbf{U}}^{H}\right)^{-1} \Phi^{H}_{d}\right) \text{Tr}\left( \left(\mathbf{K}_{\mathbf{U} \mathbf{U}}^{X}\right)^{-1} \Phi^{X}_{d}  \right).
\end{align}

\section{Additional Experiments}

\paragraph{Evaluation Metrics} To measure predictive accuracy, two evaluation metrics are considered: the normalised mean square error (NMSE) that informs on the quality of the predictive mean estimation; and the negative log predictive density (NLPD) that takes both predictive mean and predictive variance into account. 
Formally, the two metrics are defined as such:
\begin{align}
\mathrm{NMSE}= & \frac{\frac{1}{N} \sum_{i=1}^{N}\left(y_{i} - \hat{y_{i}}\right)^{2}}{\frac{1}{N} \sum_{i=1}^{N}\left(y_{i} - \bar{y}_{test}\right)^{2}},\\
\mathrm{NLPD}= & \frac{1}{2}\frac{1}{N} \sum_{i=1}^{N}\left(\left(\frac{y_{i}-\hat{y_{i}}}{\hat{\sigma}_{i}}\right)^{2}+\log \hat{\sigma}_{i}^{2}+\log 2 \pi\right),
\end{align}
where $\hat{y_{i}}$ and $\hat{\sigma}_{i}^{2}$ are respectively the predictive mean and variance for the $i$-th test point, and $y_{i}$ is the actual test value for that instance. 
The average output value for test data is $\bar{y}_{test}$.

\subsection{Simulation study: Comparing with a fixed coregionalisation matrix}
\label{compare_fixed_cor_matrix}

\begin{figure}
\centering
\includegraphics[width=0.9\textwidth]{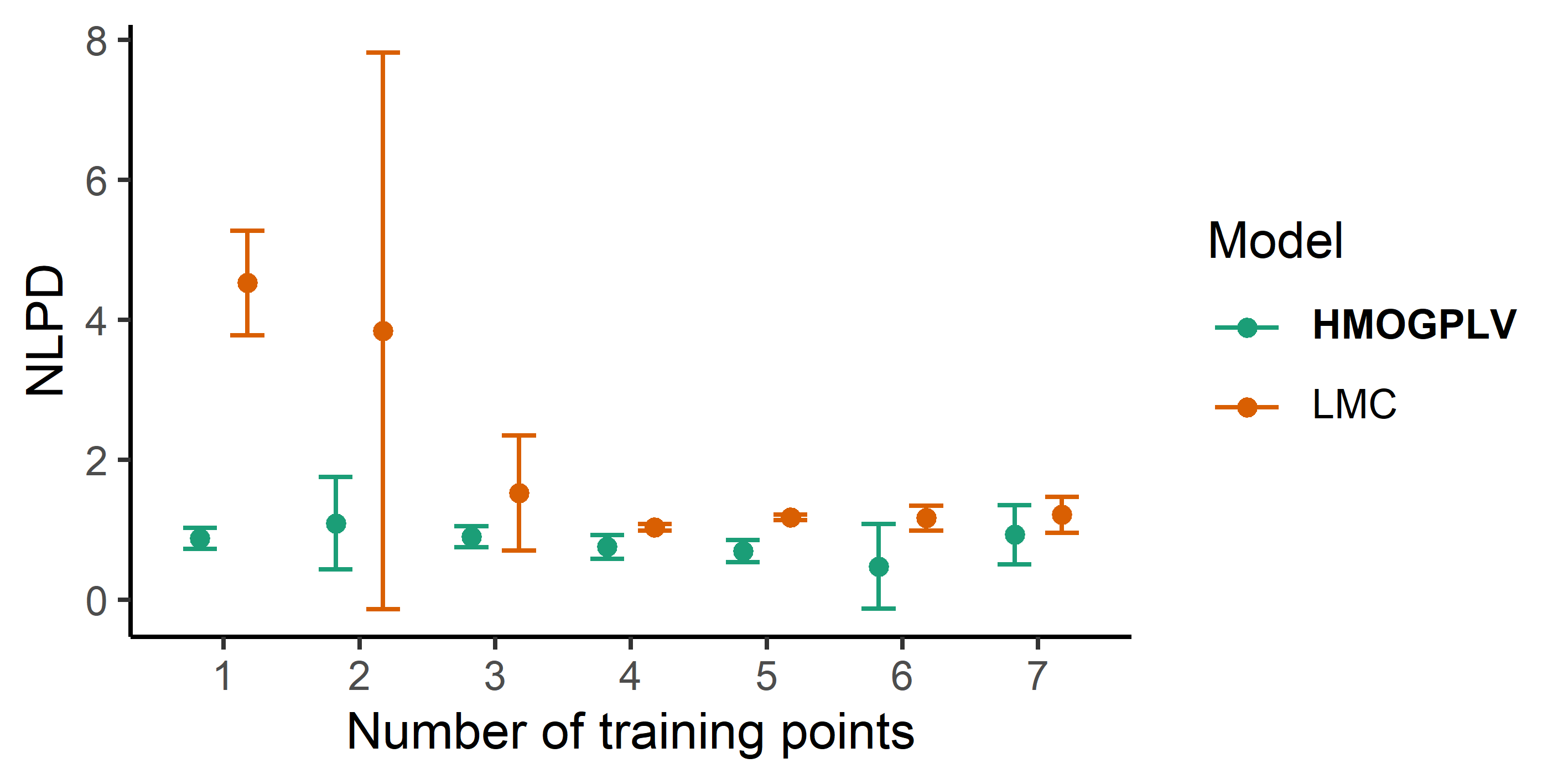}
\caption{Evolution of the prediction performance for HMOGP-LV and LMC while increasing the number of data points.}
\label{analysis_H_10_outputs}
\end{figure}

\begin{figure}
\centering
\includegraphics[width=0.9\textwidth]{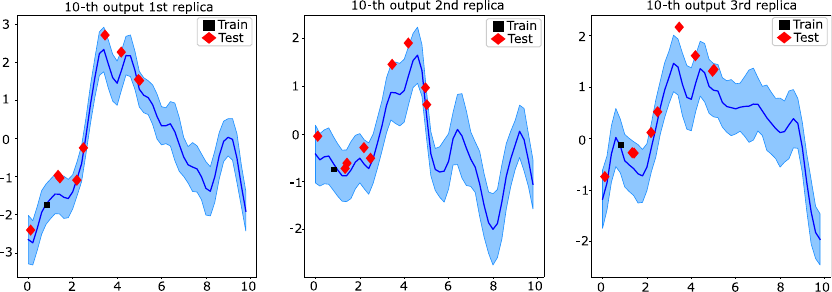}
\caption{Mean predictive curves associated with their 95\% credible intervals obtained from \textbf{HMOGP-LV} for all replicas of the testing output. The unique training point is in black, and the testing points are in red.}
\label{HMOGPLV_Using_one_data}
\end{figure}

To showcase the ability of our kernel to leverage its latent variables, we compared our method to LMC on an experiment involving 10 outputs, with 3 replicas each observed at 10 locations, where we increased the number of training points per replica sequentially and used the remaining as testing points (see Figure \ref{analysis_H_10_outputs}). This experiment does provide an intuition as to why a model based on $K_H$ can generalise better than a model based on a fixed coregionalisation matrix. An illustration of \textbf{HMOGP-LV} predictions using only one data point, depicted in Figure \ref{HMOGPLV_Using_one_data}, highlights remarkable performances for an almost-entirely missing output.
Let us mention that \textbf{HMOGP-LV} can naturally handle different input locations across outputs, which is a nice feature in many applications.

\subsection{Gene Dataset: Predicting an Entirely Missing Replica}
\begin{figure}
	\centering
	\includegraphics[width=0.9\textwidth]{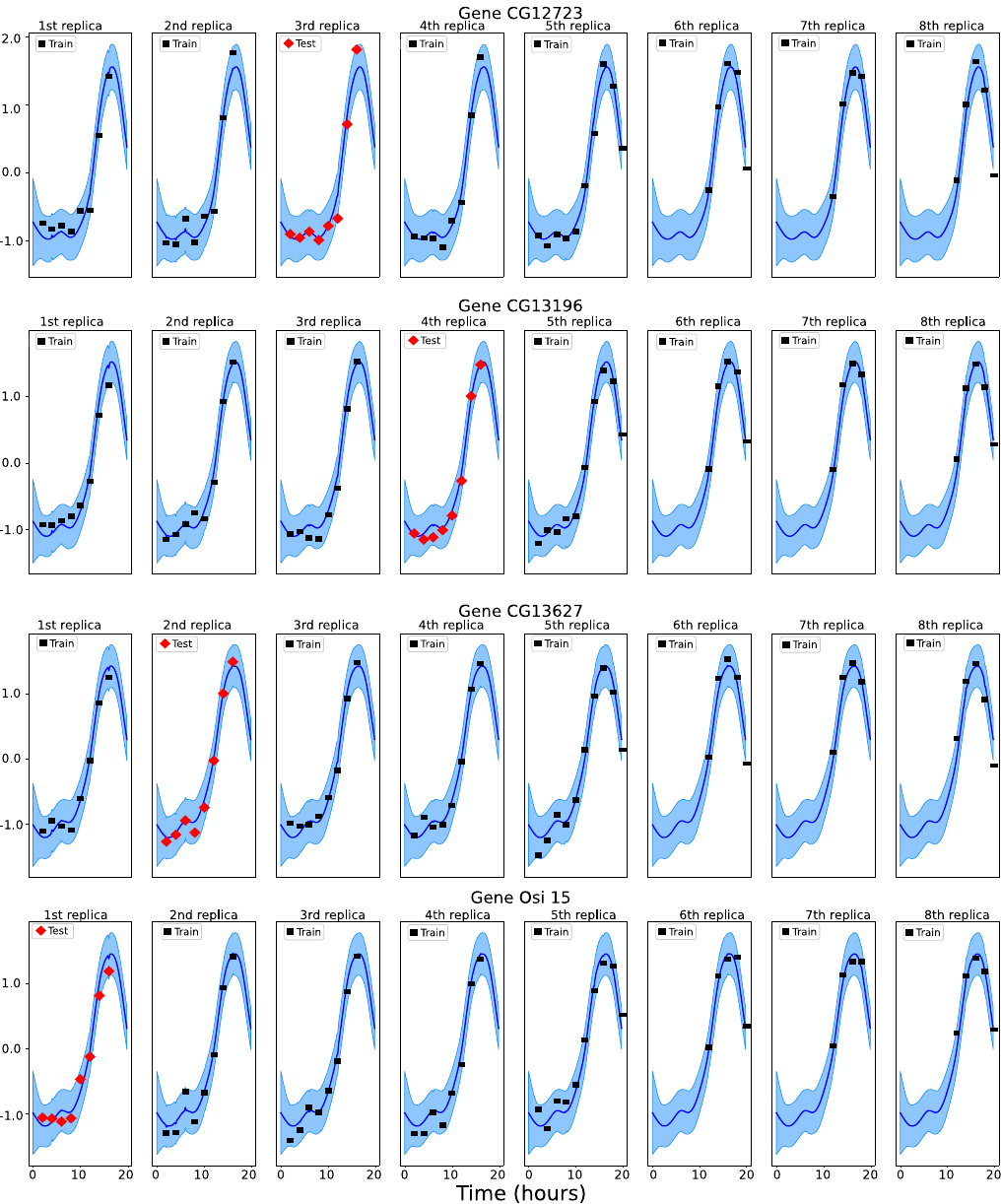}
	\caption{Mean predictive curves associated with their 95\% credible intervals for all outputs and replicas of the gene dataset. Locations of training points (in black) and testing points (in red) are specific to each output. Gene dataset with one missing replica in each output (\textbf{HMOGP-LV} performance)}
	\label{Gene-Missing-pred}
\end{figure}

\begin{figure}
	\centering
	\includegraphics[width=1\textwidth]{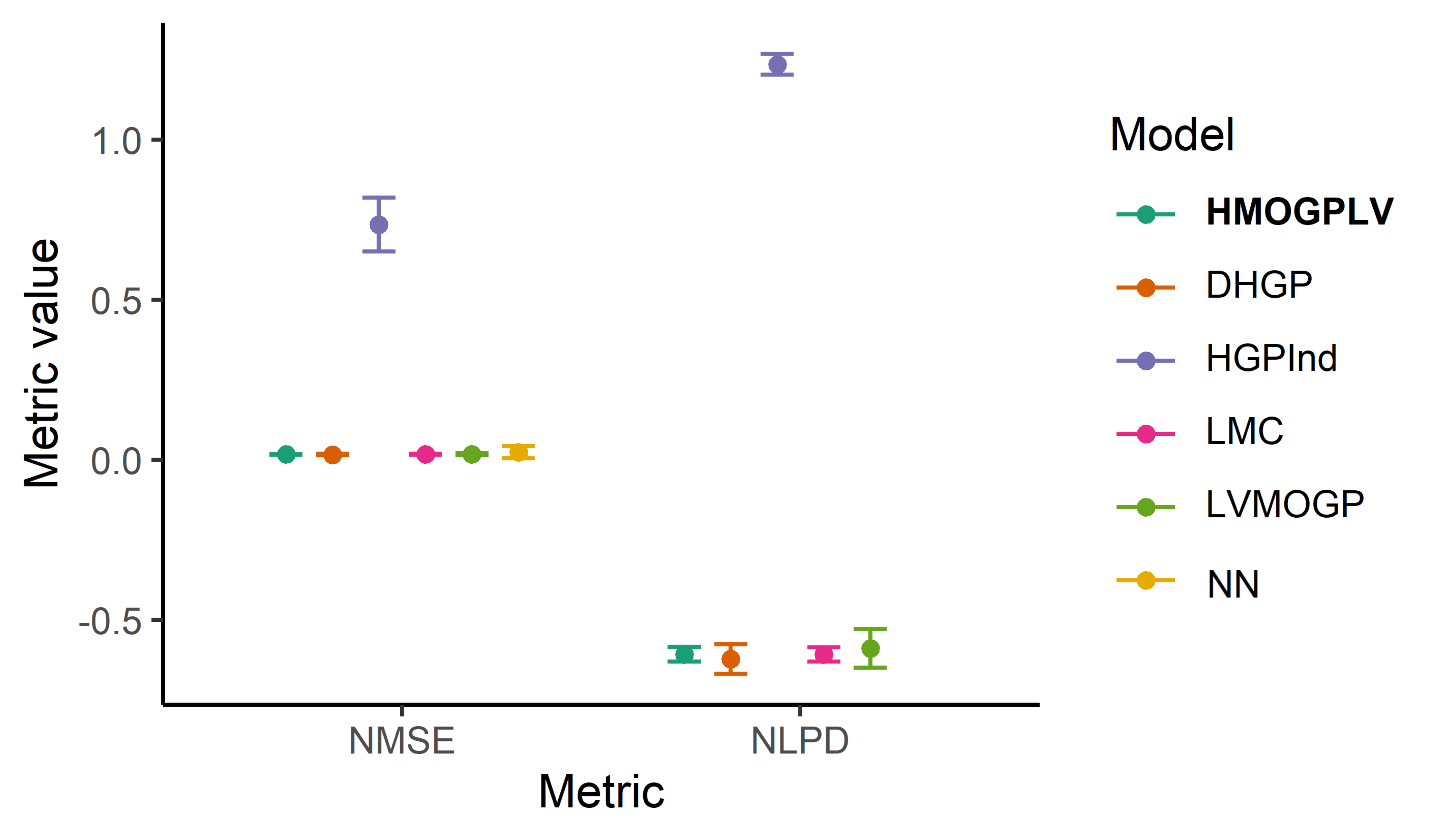}
	\caption{Gene dataset with missing one replica in each output}
	\label{Gene-Missing-replica}
\end{figure}

\begin{figure}
	\centering
	\includegraphics[width=1\textwidth]{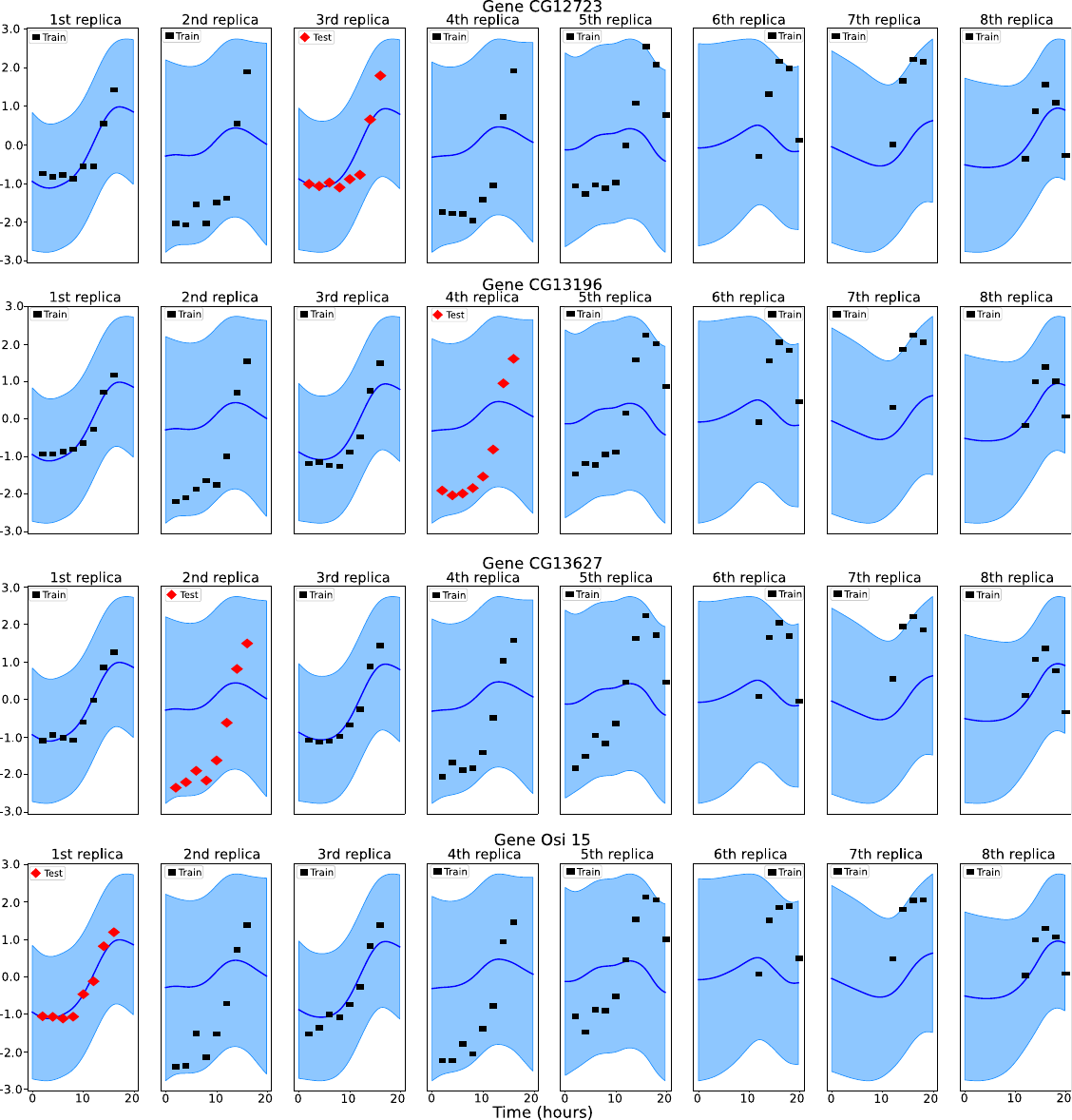}
	\caption{Mean predictive curves associated with their 95\% credible intervals for all outputs and replicas of the gene dataset. Locations of training points (in black) and testing points (in red) are specific to each output. Gene dataset with one missing replica in each output (\textbf{HGPInd} performance)}
	\label{HGPInd_gene_missing}
\end{figure}

\begin{figure}
	\centering
	\includegraphics[width=1\textwidth]{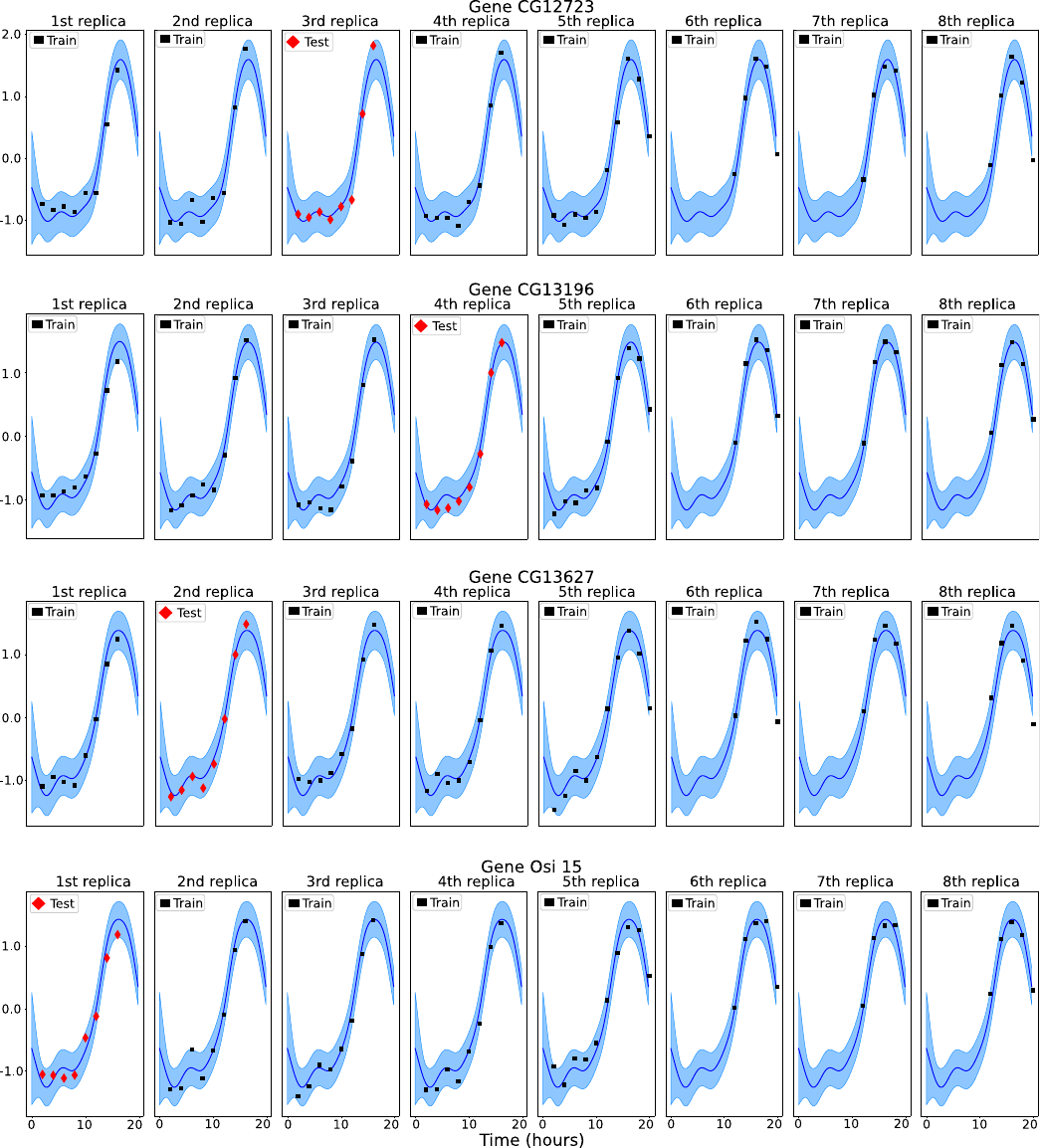}
	\caption{Mean predictive curves associated with their 95\% credible intervals for all outputs and replicas of the gene dataset. Locations of training points (in black) and testing points (in red) are specific to each output. Gene dataset with one missing replica in each output (\textbf{LMC} performance)}
	\label{LMC_gene_missing}
\end{figure}

To validate the performance of \textbf{HMOGP-LV} to handle missing replicas in real-world applications, we now apply the method on the gene dataset, where we assume there is one missing replica in each output. 
We randomly chose a missing replica per output so that seven replicas in each output are considered as training datasets.
As before, we provide in Figure \ref{Gene-Missing-pred} the visual results of \textbf{HMOGP-LV} in this experiment where one can observe that the entirely missing replicas are remarkably reconstructed with high accuracy and confidence. 
From Figure \ref{Gene-Missing-replica}, we can see that multi-output Gaussian processes approaches (e.g.  \textbf{LVMOGP} and \textbf{LMC}) also provide excellent results, comparable to our method.
In contrast, the performances of \textbf{HGPInd} appear notably poor in this context, as exhibited in Figure \ref{HGPInd_gene_missing} where it presumably captured only noise. 
We also provided the analogous visualisation for \textbf{LMC} in Figure \ref{LMC_gene_missing}.


\newpage

\subsection{Settings for the Motion Capture Dataset}
Let us provide in Table \ref{HMOGP-LV: Parameters_set_upMOCAP_data} a summary of the modelling parameter values for all experimental settings.

\begin{figure}
	\centering
\includegraphics[width=0.9\textwidth]{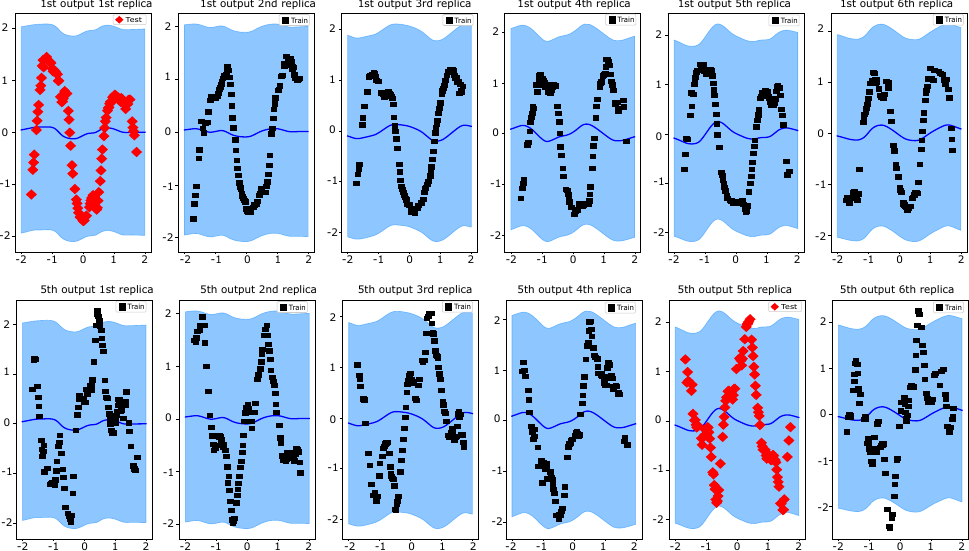}
	\caption{Mean predictive curves associated with their 95\% credible intervals for all outputs and replicas of the MOCAP-9 dataset. Locations of training points (in black) and testing points (in red) are specific to each output. (\textbf{HGPInd} performance)}
	\label{HGPInd_MOCAP9}
\end{figure}

\begin{figure}[H]
	\centering
\includegraphics[width=0.9\textwidth]{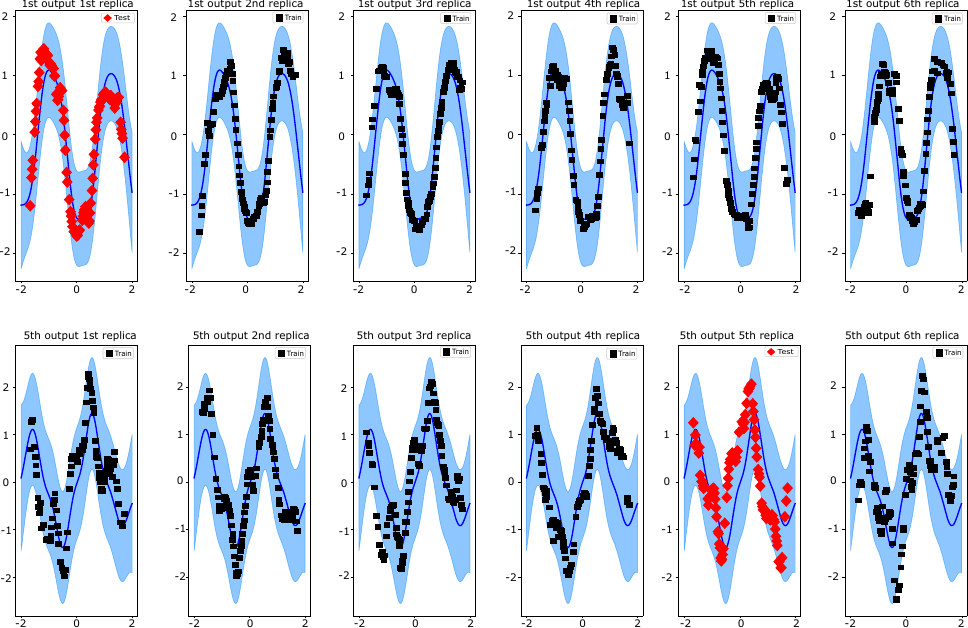}
	\caption{Mean predictive curves associated with their 95\% credible intervals for all outputs and replicas of the MOCAP-9 dataset. Locations of training points (in black) and testing points (in red) are specific to each output. (\textbf{LMC} performance)}
	\label{LMC_MOCAP9}
\end{figure}

\begin{table}
	\caption{\textsl{Setting and parameters of different GP models in MOCAP dataset. $M_{\mathbf{X}}$ indicates the number of inducing points in $\mathbf{Z}^{X}$. $M_{\mathbf{H}}$ indicates the number of inducing points in $\mathbf{Z}^{H}$. Neither \textbf{DHGP} or \textbf{NN} make use of inducing variables.}}
	\begin{center}
		\begin{tabular}{|c|c|c|c|c|c| }
			\hline
			\multicolumn{1}{|c|}{Dataset} &\multicolumn{1}{|c|}{Model} &\multicolumn{1}{|c|}{$M_{\mathbf{H}}$} &\multicolumn{1}{|c|}{$M_{\mathbf{X}}$}\\
			\hline
			\multirow{4}*{MOCAP-8}  & HMOGP-LV & \multirow{2}*{2}  & \multirow{4}*{6} \\ \cline{2-2} 
			~  &  LVMOGP & ~ & ~ \\  \cline{2-3}
			~  &  HGPInd	  & \multirow{2}*{None} & ~   \\ \cline{2-2} 
			~  & LMC	   & ~ & ~ \\ \hline
			\multirow{4}*{MOCAP-9} & HMOGP-LV  &  \multirow{2}*{5} & \multirow{4}*{5} \\ \cline{2-2}
			~ &  LVMOGP &   ~ & ~ \\ \cline{2-3}
			~   & HGPInd	 & \multirow{2}*{None}   & ~  \\ \cline{2-2} 
			~   & LMC	 & ~ & ~  \\ \hline
			\multirow{4}*{MOCAP-64} & HMOGP-LV &  \multirow{2}*{5}  & \multirow{4}*{5} \\ \cline{2-2} 
			~  &  LVMOGP & ~ & ~ \\ \cline{2-3} 
			~   & HGPInd	&  \multirow{2}*{None} & ~ \\ \cline{2-2} 
			~   & LMC	  & ~ & ~ \\ \hline
			\multirow{4}*{MOCAP-118} & HMOGP-LV   & \multirow{2}*{3} & \multirow{4}*{6} \\ \cline{2-2} 
			~ &  LVMOGP &    ~ & ~ \\ \cline{2-3}
			~   &  HGPInd	   & \multirow{2}*{None} & ~ \\  \cline{2-2} 
			~   & LMC	   & ~ & ~ \\ \hline
		\end{tabular}
	\end{center}
	\label{HMOGP-LV: Parameters_set_upMOCAP_data}
\end{table}
As for the gene dataset, we provide in Figure \ref{HGPInd_MOCAP9} and Figure \ref{LMC_MOCAP9}, the additional visualisation all predicted curves and uncertainty for both \textbf{HGPInd} and \textbf{LMC} on the MOCAP-9 dataset.

\bibliographystyle{unsrtnat} 
\bibliography{refs}

\begin{thebibliography}{15}
\providecommand{\natexlab}[1]{#1}
\providecommand{\url}[1]{\texttt{#1}}
\expandafter\ifx\csname urlstyle\endcsname\relax
  \providecommand{\doi}[1]{doi: #1}\else
  \providecommand{\doi}{doi: \begingroup \urlstyle{rm}\Url}\fi

\bibitem[Kalinka et~al.(2010)Kalinka, Varga, Gerrard, Preibisch, Corcoran,
  Jarrells, Ohler, Bergman, and Tomancak]{kalinka2010gene}
Alex~T Kalinka, Karolina~M Varga, Dave~T Gerrard, Stephan Preibisch, David~L
  Corcoran, Julia Jarrells, Uwe Ohler, Casey~M Bergman, and Pavel Tomancak.
\newblock Gene expression divergence recapitulates the developmental hourglass
  model.
\newblock \emph{Nature}, 468\penalty0 (7325):\penalty0 811--814, 2010.

\bibitem[Gelman et~al.(2013)Gelman, Carlin, Stern, Vehtari, and
  Rubin]{gelman2013bayesian}
Andrew Gelman, John~B Carlin, Hal~S Stern, Aki Vehtari, and Donald~B Rubin.
\newblock \emph{Bayesian data analysis, 3rd edition}.
\newblock Chapman and Hall/CRC, 2013.

\bibitem[Lawrence and Moore(2007)]{lawrence2007hierarchical}
Neil~D Lawrence and Andrew~J Moore.
\newblock {Hierarchical Gaussian process latent variable models}.
\newblock In \emph{Proceedings of the 24th international conference on Machine
  learning}, pages 481--488, 2007.

\bibitem[Park and Choi(2010)]{park2010hierarchical}
Sunho Park and Seungjin Choi.
\newblock {Hierarchical Gaussian process regression}.
\newblock In \emph{Proceedings of 2nd Asian conference on machine learning},
  pages 95--110. JMLR Workshop and Conference Proceedings, 2010.

\bibitem[Hensman et~al.(2013)Hensman, Lawrence, and
  Rattray]{hensman2013hierarchical}
James Hensman, Neil~D Lawrence, and Magnus Rattray.
\newblock {Hierarchical Bayesian modelling of gene expression time series
  across irregularly sampled replicates and clusters}.
\newblock \emph{BMC bioinformatics}, 14\penalty0 (1):\penalty0 252, 2013.

\bibitem[Damianou and Lawrence(2013)]{damianou2013deep}
Andreas Damianou and Neil~D Lawrence.
\newblock {Deep Gaussian processes}.
\newblock In \emph{Artificial intelligence and statistics}, pages 207--215.
  PMLR, 2013.

\bibitem[Flaxman et~al.(2015)Flaxman, Gelman, Neill, Smola, Vehtari, and
  Wilson]{flaxman2015fast}
Seth Flaxman, Andrew Gelman, Daniel Neill, Alex Smola, Aki Vehtari, and
  Andrew~Gordon Wilson.
\newblock {Fast hierarchical Gaussian processes}.
\newblock \emph{Manuscript in preparation}, 2015.

\bibitem[Li and Chen(2018)]{li2018hierarchical}
Ping Li and Songcan Chen.
\newblock {Hierarchical Gaussian processes model for multi-task learning}.
\newblock \emph{Pattern Recognition}, 74:\penalty0 134--144, 2018.

\bibitem[Dai et~al.(2017)Dai, {\'A}lvarez, and Lawrence]{dai2017efficient}
Zhenwen Dai, Mauricio~A {\'A}lvarez, and Neil~D Lawrence.
\newblock {Efficient modeling of latent information in supervised learning
  using Gaussian processes}.
\newblock \emph{arXiv preprint arXiv:1705.09862}, 2017.

\bibitem[Titsias(2009)]{titsias2009variational}
Michalis Titsias.
\newblock {Variational learning of inducing variables in sparse Gaussian
  processes}.
\newblock In \emph{\textit{Artificial Intelligence and Statistics}}, pages
  567--574, 2009.

\bibitem[Titsias and Lawrence(2010)]{titsias2010bayesian}
Michalis Titsias and Neil~D Lawrence.
\newblock {Bayesian Gaussian process latent variable model}.
\newblock In \emph{Proceedings of the Thirteenth International Conference on
  Artificial Intelligence and Statistics}, pages 844--851. JMLR Workshop and
  Conference Proceedings, 2010.

\bibitem[Goovaerts et~al.(1997)]{goovaerts1997geostatistics}
Pierre Goovaerts et~al.
\newblock \emph{Geostatistics for natural resources evaluation}.
\newblock Oxford University Press on Demand, 1997.

\bibitem[Kingma and Ba(2014)]{kingma2014adam}
Diederik~P Kingma and Jimmy Ba.
\newblock Adam: A method for stochastic optimization.
\newblock \emph{arXiv preprint arXiv:1412.6980}, 2014.

\bibitem[Virtanen et~al.(2020)Virtanen, Gommers, Oliphant, Haberland, Reddy,
  Cournapeau, Burovski, Peterson, Weckesser, Bright, et~al.]{virtanen2020scipy}
Pauli Virtanen, Ralf Gommers, Travis~E Oliphant, Matt Haberland, Tyler Reddy,
  David Cournapeau, Evgeni Burovski, Pearu Peterson, Warren Weckesser, Jonathan
  Bright, et~al.
\newblock Scipy 1.0: fundamental algorithms for scientific computing in python.
\newblock \emph{Nature methods}, 17\penalty0 (3):\penalty0 261--272, 2020.

\bibitem[Moreno-Mu{\~n}oz et~al.(2018)Moreno-Mu{\~n}oz, Art{\'e}s, and
  {\'A}lvarez]{moreno2018heterogeneous}
Pablo Moreno-Mu{\~n}oz, Antonio Art{\'e}s, and Mauricio {\'A}lvarez.
\newblock {Heterogeneous multi-output Gaussian process prediction}.
\newblock In \emph{\textit{Advances in Neural Information Processing Systems}},
  pages 6712--6721, 2018.

\end{thebibliography}

\end{document}